\documentclass[journal]{IEEEtran}
\ifCLASSINFOpdf
\else
\fi
\usepackage{framed,multirow}
\usepackage{graphicx}
\usepackage{textcomp}
\usepackage{hyperref}
\usepackage{amssymb}
\usepackage{latexsym}
\usepackage{algorithmic}
\usepackage{url}
\usepackage{xcolor}
\usepackage{cite}
\usepackage{ragged2e}
\usepackage{aligned-overset}
\usepackage[ruled,vlined]{algorithm2e}
\usepackage{bm}
\usepackage{comment}
\begin{document}
%
\title{AE-Netv2: Optimization of Image Fusion Efficiency and Network Architecture}
%
%
%
\author{\IEEEauthorblockN{Aiqing Fang,
		Xinbo Zhao,
		Jiaqi Yang,
		Beibei Qin,
	    Yanning Zhang, ~\IEEEmembership{Senior Member,~IEEE}
	 }\\

	\IEEEcompsocitemizethanks{\IEEEcompsocthanksitem 
		This work was supported by the National Natural Science Foundation of China under Grants nos. 61871326, and the Shaanxi Natural Science Basic Research Program under Grant no. 2018JM6116, and National Natural Science Foundation of China under Grants nos. 61231016. (Corresponding author: Xinbo Zhao.) The authors are with the Department of National Engineering Laboratory for Integrated Aero-Space-Ground-Ocean Big Data Application Technology, School of Computer Science, Northwestern Polytechnical University, Xi’an 710072, China. (aiqingf@mail.nwpu.edu.cn; xbozhao@nwpu.edu.cn; jqyang@nwpu.edu.cn;2019262385@mail.nwpu.edu.cn; ynZhang@nwpu.edu.cn).
			
}}

%
%

\markboth{IEEE TRANSACTIONS ON IMAGE PROCESSING, 2020}%
{Shell \MakeLowercase{\textit{et al.}}: Bare Demo of IEEEtran.cls for IEEE Journals}
%



\maketitle

\begin{abstract}
Existing image fusion methods pay few research attention to image fusion efficiency and network architecture. However, the efficiency and accuracy of image fusion has an important impact in practical applications. To solve this problem, we propose an \textit{efficient autonomous evolution image fusion method, dubed by AE-Netv2}. Different from other image fusion methods based on deep learning, AE-Netv2 is inspired by human brain cognitive mechanism. Firstly, we discuss the influence of different network architecture on image fusion quality and fusion efficiency, which provides a reference for the design of image fusion architecture. Secondly, we explore the influence of pooling layer on image fusion task and propose an image fusion method with pooling layer. Finally, we explore the commonness and characteristics of different image fusion tasks, which provides a research basis for further research on the continuous learning characteristics of human brain in the field of image fusion. Comprehensive experiments demonstrate the superiority of AE-Netv2 compared with state-of-the-art methods in different fusion tasks at a real time speed of 100+ FPS on GTX 2070. Compared with the original version of AE-Net and AE-Net after pruning, our image fusion efficiency increases 7.47 times and 2.53 times respectively. The model size of AE-Netv2 is about 69+ kb, which reduces 2808.69 times and 833.83 times respectively. Compared with the fastest image fusion method based on deep learning, the efficiency of AE-Netv2 is improved by 2.14 times. Compared with the image fusion model with the smallest model size, the size of AE-Netv2 model is reduced by 11.59 times. Compared with the image fusion method published on PAMI in 2020, the efficiency is increased by 90.21 times and the model size is reduced by 15434.20 times and the image fusion quality and continuous learning characteristics are significantly improved. Among all tested methods based on deep learning, AE-Netv2 has the faster speed, the smaller model size and the better robustness. 
\end{abstract}


%
\IEEEpeerreviewmaketitle

\section{Introduction}

\IEEEPARstart{I}{mage} fusion is widely used in many fields of computer vision. Inspired by neurobiological science, many image fusion methods based on deep learning have been proposed and achieved remarkable results. However, the \textit{robustness, generality and real-time} performance of image fusion seriously limit the application of image fusion technology. Although the first autonomous evolution image fusion method~\cite{fang2020aenet} effectively overcomes the robustness and generality of image fusion by introducing human cognitive mechanism. Unfortunately, \textit{the efficiency of image fusion is still a challenge.} However, human brain is not only robust and general for many visual processing tasks, but also has strong efficiency. 

According to the research of cognitive psychology~\cite{MillerCPortex} and biological neuroscience \cite{GuangYang2009Smds}, human brain has the ability of working memory and continuous learning~\cite{fang2019crossmodal,fang2020aenet}. In addition, Arriaga et al.~\cite{Arriaga2015Visual} pointed out that human beings can use less than 1\% of the original information to classify data and can achieve similar performance to human beings through very simple neural network architecture. The above research describe the working mechanism of human brain from a different perspective, which shows that \textit{human brain can process massive visual information without complicated modeling and massive data training}. However, in the field of image fusion, in order to improve the network performance, the depth of the network and the complexity of the model are getting higher and higher, which makes the image fusion inefficient and has a great difference compared with the human visual system. This inefficient image fusion method cannot be applied to real scenes. 1) \textbf{Unfortunately}, researchers paid few research attention to image fusion efficiency~\cite{Liu2017InfraredCNN,Liu2017MultiCNN,Yu2017medicalCNN,Yin2018MedicalNSSPAPCNN,DengXin2020DCNN,xu2020aaai,PGMI,Li2018Infrared,MaFusionGAN,fang2019crossmodale,fang2019crossmodal,ZHANG202099}. \textbf{Therefore}, we are thinking whether small sample data and simple network architecture can achieve high precision and efficient image fusion method? 
2) \textbf{In addition}, existing image fusion methods lack the application of pooling layer. This is because pooling layer leads to the loss of information, which seems to be detrimental to the image fusion task. However, pooling layer can bring more deep semantic features and more global features, which is helpful for image fusion task. Fang et al.~\cite{fang2019crossmodale} first introduce convolution feature maps with pooling layer at different depths by using pre-training weights. Li et al.~\cite{li2020nestfuse} proposed an infrared and visible image fusion method based on nest network. Pool layer operation is introduced through the nest network. Besides, there is no research and application of pooling layer in the field of image fusion. Although above methods introduce multi-scale information through pre-training model or nest network, there is no comparative analysis of different multi-scale network architectures. So, how much influence does different multi-scale network architectures have on image fusion quality? 3) \textbf{Besides}, with the gradual deepening of brain science research, image fusion methods with continuous learning ability inspired by brain gradually become the trend of image fusion. For continuous learning, Fang et al.~\cite{fang2019crossmodale,fang2019crossmodal} discussed the influence of multi-task optimization on image fusion task for the first time. In addition, Fang et al.~\cite{fang2020aenet} proposed an image fusion framework with continuous learning ability for the first time. Xu et al.~\cite{U2Fusion} proposed an unified image fusion framework for memory forgetting. The former constructs the first automatically evolving image fusion network in the field of image fusion by analyzing the continuous learning characteristics of human brain. The latter focuses on the storage of the different image fusion task features. Although above methods explore the continuous learning characteristics of image fusion, there are still some problems. For example, the influence of the commonness of different image fusion tasks on image quality.

\begin{figure}[ht]
	\centering
	\includegraphics[width=0.48\textwidth]{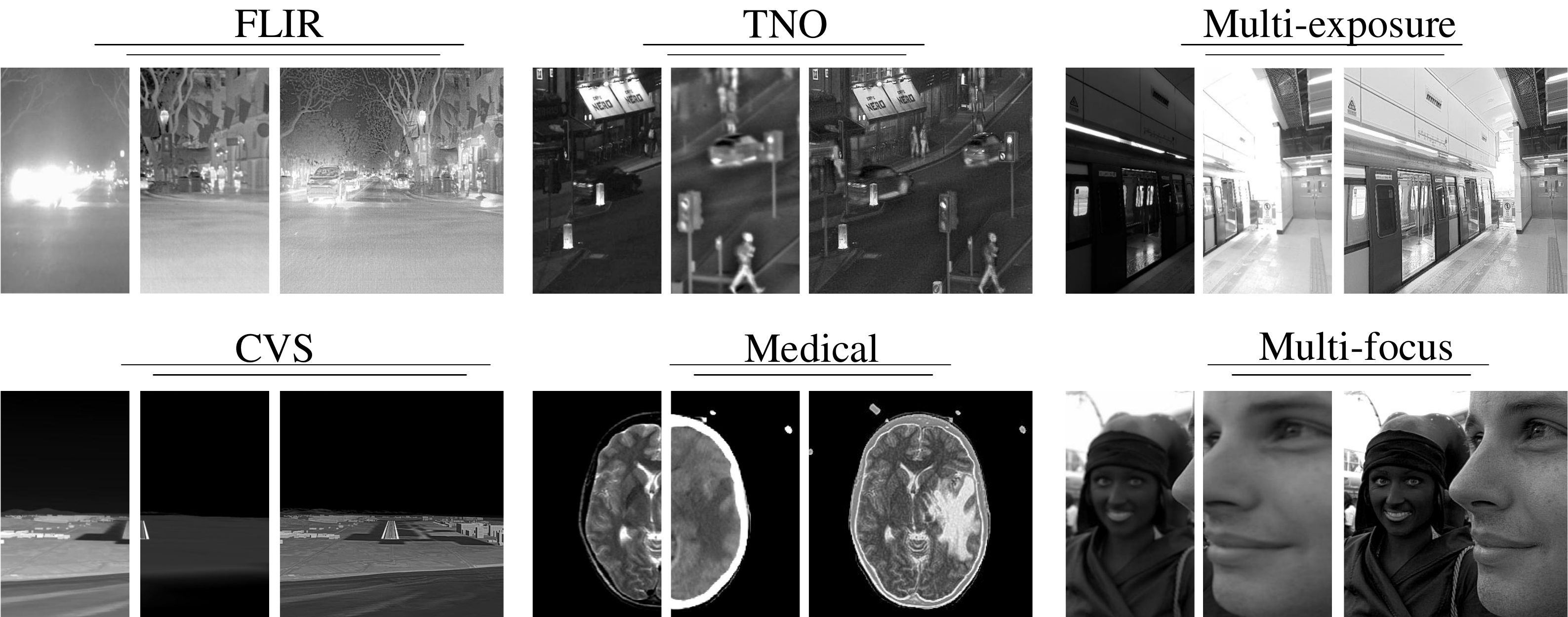}
	\caption{Examples of different image fusion datasets.\label{summary}}
\end{figure}

In order to overcome above problems, inspired by human brain cognitive mechanism, we propose an efficient, robust and general image fusion method, dubed by AE-Netv2. As far as we know, this is the first time in the field of image fusion to analyze the the robustness, generality and effiency combined with human cognitive mechanism. Our method not only guarantees the performance of image fusion, but also improves the efficiency of image fusion. Because the research idea of this paper is the same as AE-Net, the focus of this paper is to analyze the impact of above problems on image fusion task. For the description of other theories, please refer to~\cite{fang2020aenet}. The main contributions of our work include the following three points:
\begin{itemize}
	\item We explore and evaluate the influence of different network architectures on image fusion quality and efficiency. The feasibility of simple network architecture and small sample data in image fusion task is verified. As far as we know, this is the first exploration in the field of image fusion.
	
	\item We analyze the problem of multi-scale information extraction in existing deep-learning-based image fusion methods, explore the influence of pooling layer on image fusion task and propose an efficient image fusion method with pooling layer. 
	
	\item We make a research on continuous learning of human brain and explore the commonness and characteristics of different image fusion tasks. Although the image fusion effect of this method is limited by many factors, which provides a reference for the development of image fusion field.

\end{itemize}
The remainder of this paper is structured as follows. Sect.~\ref{Sec.rela} reviews relevant theory knowledge. Sect.~\ref{Sec.method} presents our exploratory research. Sect.~\ref{setup} introduces the experimental datasets, evaluation metrics and implementation details. Sect.~\ref{dis} presents a discussion and explanation. Sect.~\ref{con} gets a conclusion.

\section{Related work}
\label{Sec.rela}
\subsection{Image Fusion}
We classify image fusion methods into traditional ones and deep-learning-based ones~\cite{fang2020aenet,fang2019crossmodal,fang2019nonlinear,fang2019crossmodale}. We briefly describe these methods below.

\textit{1) Traditional ones~\cite{fang2020aenet,fang2019crossmodal,fang2019nonlinear,fang2019crossmodale}}. 
Traditional image fusion methods mainly include multi-scale transformation~\cite{Zhao2017Multisensor,Cui2015Detail,ZhangInfrared,Bhatnagar} and visual significance~\cite{Bavirisetti2016Two,Zhang2015A,Liu2017InfraredJSR-SD,Ma2017InfraredWLS,Cui2015Detail,ZhangInfrared} et al. Traditional image fusion methods use artificial design algorithm, so it adds human subjective intention to a certain extent, which makes the algorithm more suitable for specific image fusion task. However, compared with the deep learning methods, these methods have the following problems: firstly, the feature extraction is not sufficient and the feature extraction operator needs to be designed by feature engineering; secondly, the generality is poor and it is difficult to extend to multiple image fusion tasks~\cite{fang2019crossmodale}. Considering that the focus of this paper is image fusion method based on deep learning, you can refer to~\cite{fang2019crossmodale,fang2019crossmodal,fang2019nonlinear,fang2020aenet} for the progress of traditional image fusion methods in this part.

\textit{2) Deep-learning-based ones~\cite{fang2020aenet,fang2019crossmodal,fang2019nonlinear,fang2019crossmodale,LIU202071}}. 
For multi-focus image fusion tasks, Liu et al.~\cite{Liu2017MultiCNN} first proposed a multi-focus image fusion method with a deep convolutional neural network based on binary classification. For multi-exposure image fusion tasks, Prabhakar et al.~\cite{PrabhakarK.Ram2017DADU} proposed the first deep unsupervised approach for exposure fusion with extreme exposure image pairs. For infrared and visible image fusion tasks, Liu et al.~\cite{Liu2017InfraredCNN} proposed the first infrared and visible image fusion method with convolutional neural networks. For medical image fusion tasks, wang et al.~\cite{MPCNN} proposed the first medical image fusion method using m-PCNN. For remote sensing image fusion tasks, Masi et al.~\cite{GiuseppePansharpening} proposed the first pansharpening method with convolutional neural networks. For combined vision system image fusion tasks, Fang et al.~\cite{fang2019crossmodale} proposed the first CVS image fusion method by subjective visual attention. Above methods combined with the specific field of image fusion to explore the convolution neural network method for the first time. For the problem of robustness and generality of image fusion tasks, Zhang et al.~\cite{ZHANG202099} proposed the first general image fusion framework based on convolutional neural network. Although the method needs to choose different fusion rules for different image fusion tasks, the robustness and generality of the method has been greatly improved compared with previous work. In addition, Deng et al.~\cite{DengXin2020DCNN} proposed a novel deep convolutional neural network to solve the general multi-modal image restoration and multi-modal image fusion problems. Ma et al.~\cite{PGMI} proposed a fast and general image fusion network to overcome the problem of information loss through feature reuse.
Although the network architecture of above methods can be applied to multiple image fusion tasks, the strategy of migration learning is adopted to adapt to the differences of different data distribution. This will lead to the problem of memory forgetting, which makes the transferred model cannot be directly used in the previous data distribution. Recently, Xu et al.~\cite{xu2020aaai, U2Fusion} proposed an elastic weight consolidation method to avoid forgetting problem. For the lack of research on human visual cognition mechanism, fang et al.~\cite{fang2019crossmodale,fang2019crossmodal} proposed the first image fusion method of subjective attention and multi-task learning. In addition, Fang et al.~\cite{fang2020aenet} proposed the first autonomous evolution image fusion method with continuous learning ability inspired by human brain cognitive mechanism. As far as we know, this is the first image fusion network architecture with autonomous evolution and continuous learning ability. Of course, there are many other image fusion methods based on deep learning, such as FusionGan~\cite{MaFusionGAN}, DDcGAN~\cite{MA202085}, DenseFuse~\cite{Li2018DenseFuse} et al. Above methods are all end-to-end deep learning image fusion methods, compared with the traditional image fusion methods in robustness and generality has achieved remarkable results. However, researches pay few research attention to the relationship between efficiency, robustness, generality and human brain characteristic.
After all, the human brain is not only efficient in processing tasks, but also robust. and can quickly adapt to complete a variety of tasks. After all, the human brain has high efficiency, robustness and generality in many visual processing tasks.

\subsection{Network Architecture}
We mainly describe the related work of image fusion network architectures based on deep learning. 

The network architecture of image fusion method based on deep learning includes twin network and generative adversarial network. The image fusion framework based on twin network mainly uses two-way branch network or one common branch network to extract the features of the image to be fused. The feature maps are aggregated and fused by different fusion criterias~\cite{PrabhakarK.Ram2017DADU,Liu2017MultiCNN,Liu2017InfraredCNN,MPCNN,GiuseppePansharpening,fang2019crossmodale,ZHANG202099,DengXin2020DCNN,PGMI,xu2020aaai,fang2020aenet,MA202085,Li2018DenseFuse}. In the framework of image fusion based on generative adversarial network, the weight of image fusion is optimized by generative adversarial loss function. Ma et al.~\cite{MaFusionGAN} proposed the generative adversarial loss in infrared and visible image fusion task. In order to solve the problem of fuzzy image fusion and loss of fusion information, Ma et al.~\cite{MA202085} proposed an infrared and visible image fusion method via detail preserving adversarial learning. In addition, Ma et al.~\cite{PANGAN} introduced the generative adversarial loss into remote sensing image fusion task. However, the efficiency and architecture of deep neural networks have been studied in other computer vision processing tasks, Such as SqueezeNet~\cite{IandolaForrest2016SAaw}, Deep Compression~\cite{Han2016Deep}, Distilling~\cite{HintonGeoffrey2015DtKi}, XNorNet~\cite{RastegariMohammad2016XICU} and ShuffleNet~\cite{ZhangXiangyu2017SAEE} et al. Above methods are mainly tried from two aspects of network structure design and weight deep compression. The experiment has achieved remarkable results and greatly improved the efficiency of deep network model. Unfortunately, existing image fusion methods have paid few research attention to image fusion efficiency and network architecture.

In conclusion, inspired by the human brain task-oriented processing mechanism~\cite{fang2020aenet}, we propose an efficient, robust and general image fusion method and design a network with excellent performance, which makes the network model size the smallest and the image fusion speed the fastest without reducing the performance of AE-Net~\cite{fang2020aenet}. We first analyze the efficiency, network architecture and accuracy of deep learning methods in the field of image fusion, which is of great significance to promote the development of image fusion.

\section{Method}
\label{Sec.method}
The content of this paper is structured as follows. 1) This paper mainly explores the impact of image fusion architecture on the efficiency and accuracy of image fusion and provides reference for the design of network architecture in the field of image fusion. 2) The influence of pooling layer on the accuracy of image fusion is explored, which fills the gap in the field of image fusion. 3) In addition, we further explore the characteristics and common problems of continuous learning ability on the basis of AE-Netv2, which provides a reference for further improving the autonomous evolution ability of image fusion.

\begin{figure}[ht]
	
	
	\includegraphics[scale=1,width=0.43\textwidth]{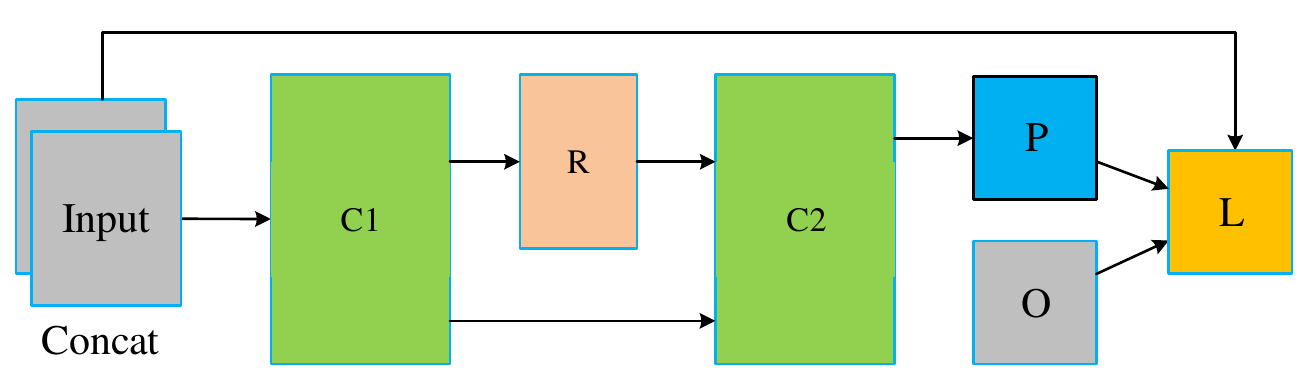}
	\centering
	
	\caption{Image fusion network architecture of AE-Netv2, where $C_1$ and $C_2$ indicate convolution block, which includes convolution, batch normalization and relu operations; $R$ represents the replaceable convolution modules include group convolution, compressed convolution, depth separable convolution, conventional convolution and some inception modules; $P$ indicates the output of image fusion network; $O$ represent optimal result of fused image; $L$ represents loss function. }
	\label{f2}
\end{figure}

\begin{algorithm}[t]
	\caption{Method of AE-Netv2 \cite{fang2020aenet}.}
	
	\label{alg:algorithm2}
	\KwIn{Input images: $x_i,y_i$; \\
		Fusion algorithm module: $F_i$; \\
		Quality evaluation module: $E_i$; \\
		Loss function: $L_p(\bf x,y)$ \\
		Fused weight: $\alpha, \beta$.} 
	\KwOut{Predict image ${\cal P}_i$}  
	\BlankLine
	
	Compute $F_i(\bf x_i,y_i)$; \\
	\For{$i = 0 \to n-1$;}{
		$F_1 = M(\bf x_i,y_i)$;	\\
		$F_2 = C(\bf x_i,y_i)$;  \\
		$... = ...$;	\\
		$F_n= I(\bf x_i,y_i)$;	\\
		$F_i(x,y)=\{ F_1, F_2, ... , F_n \}$;}
	
	Compute optimal result $O_i=E_i(F_i(\bf x,y))$;\\
	\For{$i = 0 \to n-1$;}{
		$O_1 = E_i(\bf F_1)$;	\\
		$O_2 = E_i(\bf F_2)$;  \\
		$... = ...$;	\\
		$O_n=  E_i(\bf F_n)$;	\\
		$O_i(x,y)=\{O_1, O_2, ... , O_n \}$;}
	
	\If{$O_{new} \not\in O$}
	{ $O=E_i(O_i,O_{new})$;}
	
	\While{\textnormal{loss$>t$}}{
		
		\textit{\textbf{Compute}} ${\cal P}_i  = \gamma*(\alpha*Concat(\bf x_i,y_i)*(1+\beta))$; \\
		Compute $L_p(\bf x,y)$; \\
		\If{Supervised}{ $L_p(\bf x_i,y_i)=Argmin({E}(\bf {x_i}, {p_i})+{E}(\bf {y_i}, {p_i})) / 2.0 $;}
		\Else
		{
			$L_p(\bf x_i,y_i)=Argmin{E}({p_i}, {O})+C[L(\bf x_i,y_i),0]$;	
		}
		
		Update $\alpha$ and $\beta$; 	
	}
	
\end{algorithm}

\subsection{Motivation of AE-Netv2}
In the field of computer vision, the evaluation of a model is not based on a single accuracy metrics, but also needs to consider the generality, robustness and efficiency of the model. We believe that this viewpoint is also applicable to the field of image fusion. In addition, human brain has robustness, generality and high efficiency in a variety of computer vision processing tasks. Inspired by above viewpoints, we believe that image fusion task should also have the following three characteristics: robustness, generality and efficiency. Unfortunately, existing image fusion methods from the traditional LP to the latest U2Fusion paid few research attention to image fusion efficiency, network architecture and the characteristic of human brain. However, the model size and fusion efficiency are of great significance to the promotion and application of image fusion technology. 

\subsection{Image Fusion Modeling}
In order to improve the robustness and generality of image fusion technology, existing image fusion methods usually adopt transfer learning method or multi-task learning strategy. For transfer learning, on the one hand, the pre-trained weight is used as the backbone network to extract deep semantics features. On the other hand, the original domain weight is transferred to the target domain to fine tune the weight. However, this method makes the network model with high complexity and high computational cost. In addition, this method introduces a serious memory forgetting problem, which makes the migrated model lose the fusion performance of the original domain. The strategy for multi-task is to use multi-task assisted learning mechanism to mine the hidden features between different tasks, so as to improve the robustness and generality of image fusion. However, the network complexity of this method is very high and it has high requirements for the design of network architecture, otherwise harmful features will be introduced. To solve these problems, we propose an efficient, robust and general image fusion network architecture inspired by brain cognitive mechanism. In order to ensure the efficiency, robustness and generality of image fusion network, on the one hand, we simplify the network depth and only retain several layers of convolution. On the other hand, we introduce group convolution and channel shuffling to simplify the network complexity. The reason why we remove too many convolution layers is that we find traditional convolution operation will introduce very large space complexity, which makes the model size increase rapidly in the experimental process. The introduction of group convolution can reduce the computational complexity and the internal communication mechanism of group convolution can be improved through channel shuffling operation. 

Suppose the input images $x_i$ and $y_i$, the fused image ${\cal P}_i$ is defined as:
\begin{equation}
{\cal P}_i  = \gamma*(\alpha*Concat(\bf x_i,y_i)*(1+\beta)),
\end{equation}
where i represents the i-th image; $\alpha$ and $\beta$ represent the fusion weights of the first level convolution and the group convolution are represented respectively; $\gamma$ indicates the convolution weight of the last layer. This change is the biggest difference between AE-Netv2 and AE-Net. Black italics is the key of this paper. In the algorithm \ref{alg:algorithm2}, $M(\bf x_i,y_i)$, $C(\bf x_i,y_i)$, $I(\bf x_i,y_i)$ represent multi-exposure, combined visual image and infrared and visible image fusion algorithm, respectively; $O$ indicates optimal solution.

\subsection{Image Fusion Architecture with Pooling Layer}
\begin{figure}[ht]
	
	
	\includegraphics[scale=1,width=0.45\textwidth]{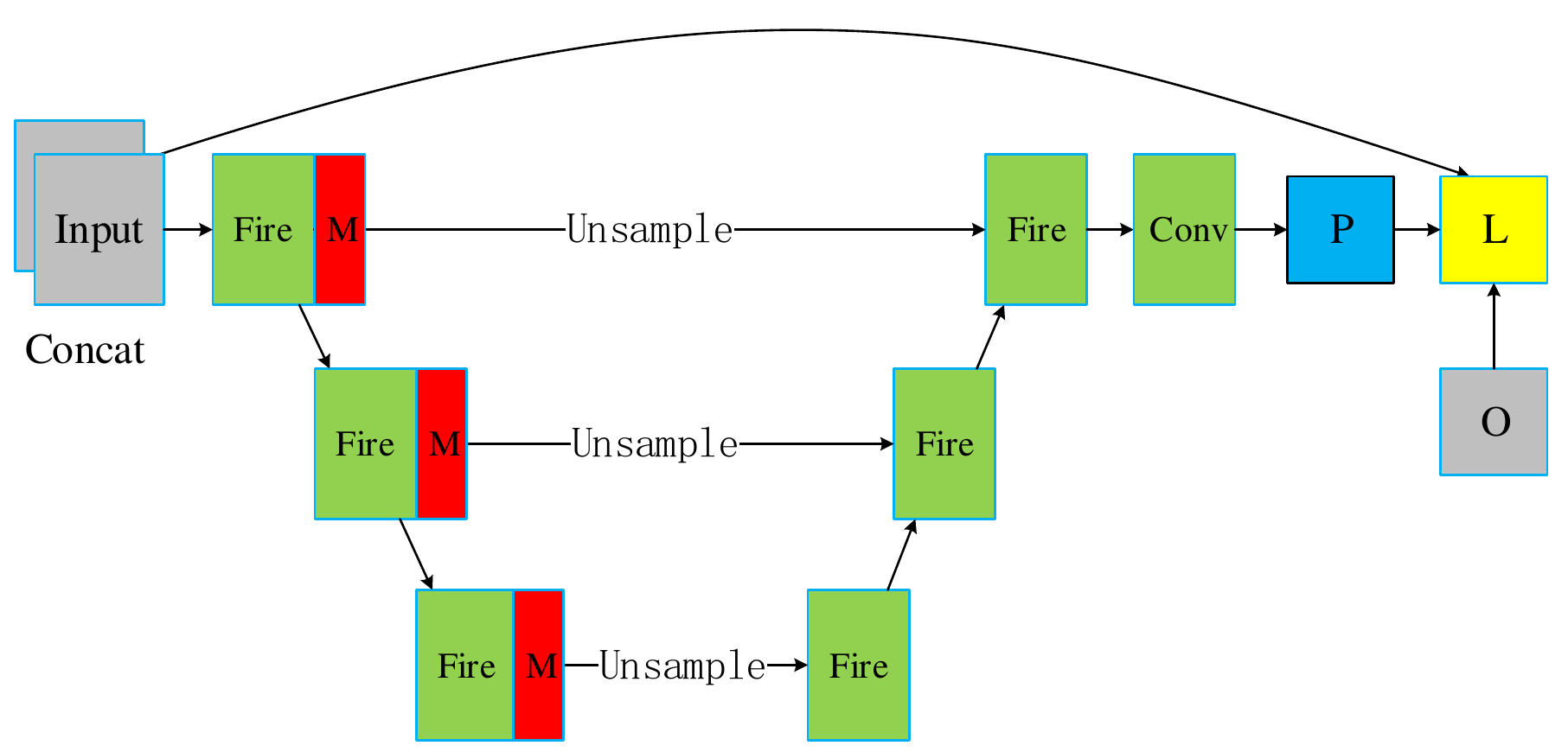}
	\centering
	\caption{Image fusion network architecture with pooling layer. Where Red box M represents pooling layer; green box fire represents squeezenet network module; $P$ represents prediction image; $L$ represents loss function; $O$ represents relative optimal solution.}
	\label{f3}
\end{figure}

Pooling layer has been widely used in many fields of computer vision and achieved remarkable results. However, in the field of image fusion, there is little research in this area. In this paper, we explore and propose an image fusion method with pooling layer based on the fire layer of squeeze and U-Net architecture. Because the pooling layer is added to the image fusion network, the network architecture of AE-Netv2 is transformed from Fig.~\ref{f2} to Fig.~\ref{f3}.

Compared with existing image fusion networks, our network effectively integrates the advantage of continuous learning ability, multi-scale features and squeeze convolution, which makes the network model very efficient and robust. Although pooling can lead to the loss of information, due to the U-Net architecture, the details of the underlying texture can be effectively preserved through hop connection.

\subsection{Commonness and Characteristics of Image Fusion}
Image fusion can be seen as a combination of common features and unique features~\cite{DengXin2020DCNN}.  In the task of image fusion based on deep learning, the image fusion quality is highly dependent on ground truth labels and loss functions. In addition, transfer learning can make the model suitable for different image fusion tasks, it will cause the problem of feature forgetting~\cite{U2Fusion}. Although above methods have achieved remarkable results, there is still the problem of losing effective features when multiple image fusion tasks are involved. Therefore, starting from the commonness and characteristics of image fusion, we construct the characteristics of different image fusion tasks on the basis of extracting common features from AE-Netv2. The difference between our method and CU-Net~\cite{DengXin2020DCNN} is that the latter explores the commonness and characteristics feature maps between two images to be fused, while what we explore is the commonness and characteristics fusion weight between different image fusion tasks. \textit{We are thinking about whether the combination of commonness and characteristics of different tasks can achieve better results like~\cite{DengXin2020DCNN}$?$} In view of this problem, we have carried on the exploratory research analysis.

Suppose the input images $x_i$ and $y_i$. For any image fusion task $M_{pi}$, if the mixed data training model is used, then $M_{pi}$ is defined as:
\begin{equation}
M_{pi}=\underbrace{M(\bf x,y),C(\bf x,y),...,I(\bf x,y)}_{\text {common fusion weight}},
\end{equation}

For the image fusion method using transfer learning, $M_p$ is defined as:
\begin{equation}
M_{pi}=\underbrace{U_t(\bf x,y)}_{\text {unique fusion weight }}+\underbrace{\alpha \times U_s(\bf x,y)}_{\text {common fusion weight }},
\label{E3}
\end{equation}
where $U_t(\bf x,y)$ represents target domain unique image fusion weight; $U_s(\bf x,y)$ represents source domain unique image fusion weight; $\alpha$ indicates the weight coefficient of original domain image fusion. Accoding to the principle of CU-Net, $M_{pi}$ is defined as:
\begin{equation}
M_{pi}=\underbrace{U_i(\bf x,y)}_{\text {unique fusion weight }}+ \underbrace{M(\bf x,y),C(\bf x,y),...,I(\bf x,y)}_{\text {common fusion weight }},
\end{equation}
where $U_i(\bf x, y)$ represents any image fusion task unique features. For the same image fusion task, this is correct. However, is it really right to directly explore the commonalities and characteristics of multiple tasks? In the experiment, we find that the common fusion weight obtained by AE-Netv2 is better than the feature fusion weight of single task. Therefore, it would be more appropriate to define $M_{pi}$ as follows:
\begin{equation}
M_{pi}=\underbrace{U_i(\bf x,y)}_{\text {unique fusion weight }}+ \beta \times \underbrace{M(\bf x,y),C(\bf x,y),...,I(\bf x,y)}_{\text {common fusion weight }},
\label{E5}
\end{equation}
where $\beta$ represent the degradation coefficient of common fusion weight. In this case, Eq.\ref{E5} seems to be similar to transfer learning Eq.\ref{E3}, but it is not the same. The former does not necessarily lead to the degradation of image fusion quality. Image quality is directly related to loss functions and ground truth labels. The latter will lead to the degradation of image fusion quality. This is because the image fusion weight is no longer like the intersection of CU-Net but a more complex promotion effect. However, this phenomenon is not suitable for the traditional image fusion method based on similarity loss function. This is because the traditional loss of structural similarity or MSE learn more about the similarity of the original image, rather than improving the image quality.

\subsection{Image Fusion Network Architecture and Traning Skills}
As shown in Fig.~\ref{f2}, we can find that the AE-Netv2 network architecture is very simple, only including two traditional convolution blocks, one group convolution and one channel shuffling operation. This is why our method can be very efficient. \textit{Why do we combine traditional convolution with group convolution$?$} Here we need to point out that GCB is a replaceable module to explore the balance between the efficiency and quality of image fusion. So, is the combination of regular convolution and group convolution the best choice$?$ No. In the last experimental section, we will compare the regular convolution, group convolution, separate convolution, squeeze module, inception module and combined module. In order to speed up the convergence speed of the model and improve the training efficiency, we adopt the method of segment training. First of all, 10 epochs are trained with a larger learning rate of 0.001 which can quickly converge to 0.4. Then, the weight of the model is saved and the learning rate is 0.0001 to continue training 100 epoches which can ensure that the training of the network model can be completed in 30 minutes. Our network model parameters are shown in Table~\ref{table111} and network hyperparameters are shown in Table~\ref{table222}.

\begin{table}[!h]
	
	\centering \footnotesize
	\renewcommand \arraystretch{1.1}
	\caption{
		Network Architecture Parameters}
	\label{table111}
	\begin{tabular}[b]{p{1cm}p{1cm}p{1cm}p{1.5cm}p{1.5cm}}
		\hline
			Phase   &   Input & Output & Kernel & Image size\\ \hline
			C1 &   6 & 64&3&128$\times$128\\ 
			GC &    64& 64&3&128$\times$128	\\
			C2 &   64 & 1&3&128$\times$128\\ 
			\hline	
	\end{tabular}

\end{table}

\begin{table}[!h]
	
	\centering \footnotesize
	\renewcommand \arraystretch{1.1}
	\caption{
		Network hyperparameters
	}
	\label{table222}
	\begin{tabular}[b]{p{1cm}p{1cm}p{1cm}p{2cm}p{1.5cm}}
		\hline
		Phase   &   Batchsize & Epoach & Learning rate & Image size\\ \hline
		First &  16  & 10&0.001&128$\times$128\\ 
		Second &  16  & 100&0.0001&128$\times$128\\ \hline	
	\end{tabular}

\end{table}

\section{Experiments}
\label{setup}
In this section, experimental setup are first presented. Then, comparative experiment and analysis experiment are introduced \cite{fang2020aenet}.
\begin{table*}[!h]
	
	\centering \footnotesize
	\renewcommand \arraystretch{1.1}
	\caption{Metrics \cite{fang2020aenet}}
	\label{table122}
	\begin{tabular}[b]{p{0.05cm}p{1.6cm}p{7cm}p{7cm}}
		\hline
		No.& Method &Equations   & Description \\ 
		\hline
		1&EN \cite{1576816}&$EN=-\sum_{i=0}^{255} p_{i} \log _{2} p_{i},$&where $P_i$ is the probability of a gray level appearing in the image.\\ 
		\hline
		2&AG \cite{Cui2015Detail}&$AG=\frac{1}{M^{*} N} \sum_{i=1}^{M} \sum_{j=1}^{N} \sqrt{\frac{\Delta I_{x}^{2}(\bf i, j)+\Delta I_{y}^{2}(\bf i, j)}{2}},$& where $M \times N$ denotes the image height and width; $\Delta I_{x}(\bf i, j)$ denotes image horizontal gradient; $\Delta I_{y}(\bf i, j)$ denotes image vertical gradient.\\ 
		\hline
		3&SSIM \cite{1284395}& $\begin{array}{l}
		{SSIM}(\bf I_i, R)=\frac{\left(2 u_{I_i} u_{R}+C_{1}\right)\left(2 \sigma_{I_i R}+C_{2}\right)}{\left(u_{\bf I_i}^{2}+u_{R}^{2}+C_{1}\right)\left(\sigma_{I_i}^{2}+\sigma_{R}^{2}+C_{2}\right)},\end{array}$&where $\mu_{I_i}$ and $\mu_R$ indicate the mean value of origin image $I_i$ and fused image $R$; $\sigma_{I_iR}$ is the standard covariance correlation.\\ 
		\hline
		4&VIFF \cite{Han2013A}& $\mathrm{VIF}=\frac{\sum_{j \in  { subbands }} I\left({C} \stackrel{N, j}{;} {F}^{N, j} | s^{N, j}\right)}{\sum_{j \in \ { subbands }} I\left({C}^{N, j} ; {E}^{N, j} | s^{N, j}\right)},$& where ${C} \stackrel{N, j}{;}$ denotes N elements of the $C_j$ that describes the coefficients from subband j; $\sum_{j \in  { subbands }} I\left({C} \stackrel{N, j}{;} {F}^{N, j} | s^{N, j}\right)$ denotes reference image information.\\  
		
		\hline
		5&NIQE \cite{MittalA2013MaCB}& $\begin{array}{l}
		D\left(\nu_{1}, \nu_{2}, \Sigma_{1}, \Sigma_{2}\right) \\ \quad=\sqrt{\left(\left(\nu_{1}-\nu_{2}\right)^{T}\left(\frac{\Sigma_{1}+\Sigma_{2}}{2}\right)^{-1}\left(\nu_{1}-\nu_{2}\right)\right)}\end{array},$&where $\nu_{1}$ and $\Sigma_{1}$ are the mean vectors and covariance matrices of the natural MVG model and the distorted image’s MVG model.\\  
		
		\hline
		
		6&PSNR \cite{SijbersJ1996Qaio} &$\mathrm{PSNR}=10 \log _{10} \frac{\left(2^{n}-1\right)^{2}}{M S E},$& where $MSE$ is the mean square error of the current image X and the reference image y.\\ 
		
		\hline
		
		7&MI   \cite{Qu2002Information}&$\mathrm{MI(I_i,R)}=H(I_i)+H(R)-H(I_i, R),$&
		where $H(I_i)$ and $H(R)$ represent the information entropy of origin image and fused image; $H(\bf I_i,R)$ denotes joint information entropy.\\   
		
		\hline
		8&Combined \cite{fang2020aenet}&$E_2 $ \cite{fang2020aenet}&	$E_2$ represents a variety of indicators to jointly represent image quality. For specific definition.\\ 
		\hline
	\end{tabular}
\end{table*}
\begin{table}[!h]
	
	\centering \footnotesize
	\renewcommand \arraystretch{1.1}
	\caption{
		Experimental datasets and inherited properties \cite{fang2020aenet}
	}
	\label{table11}
	\begin{tabular}[b]{p{2cm}p{1cm}p{2cm}p{0.5cm}p{1cm}}
		\hline
		Type & Dataset   & Modality & Align & Matching pairs\\ \hline
		Multi-exposure \cite{JianruiCai2018LaDS}  &   \cite{JianruiCai2018LaDS} &  Multi-exposure  &  \checkmark&4, 413\\ 
		Multi-focus \cite{Nejati2015Multi, ZhangXingchen2020MIFA}  & Lytro and MFIF   &Multi-focus  &\checkmark& 125 \\ 
		Medical \cite{Summers2003Harvard}  & Brain    & CT, MRI  &\checkmark&97    \\ 
		Infrared and visible \cite{zhang2020vifb, FLIR, LiChenglong2019RotB}  & FLIR, VIFB and RGBT  & Infrared and visible  &  \checkmark& 14, 473  \\ 
		
		Combined vision system images (OURS) & CVS  & Enhanced and synthetic vision images  &  $\times$&4, 000  \\ 
		\hline
	\end{tabular}
\end{table}

\subsection{Experimental Setup}
In this section, datasets, metrics and tested methods for experimental evaluation are first presented. Then, implementation details of evaluated methods are introduced~\cite{fang2020aenet}. 

\textit{1) Datasets:} 
\textit{a) Multi-exposure}~\cite{JianruiCai2018LaDS}
\textit{b) Multi-focus}~\cite{Nejati2015Multi}
\textit{c) Medical}~\cite{Summers2003Harvard}
\textit{d) Infrared and visible}~\cite{zhang2020vifb,LiChenglong2019RotB, FLIR}
\textit{e) Combined vision system}~\cite{Summers2003Harvard}. The 7 datasets in the experiment involve 5 image fusion tasks. Since the dataset used is the same as AE-Net, please refer to the citation or~\cite{fang2020aenet} for details of datasets. The main properties of experimental datasets are summarized in Table.~\ref{table11}~\cite{fang2020aenet}.

\textit{2) Metrics~\cite{fang2020aenet}:} 
Entropy (EN)~\cite{1576816}, average gradient (AG)~\cite{Cui2015Detail}, structural similarity (SSIM)~\cite{1284395}, visual information fidelity (VIF)~\cite{Han2013A}, natural image quality evaluator  (NIQE)~\cite{MittalA2013MaCB}, PSNR and $E_2$~\cite{fang2020aenet}. Specific attributes of metrics can refer to \ref{table122} and~\cite{fang2020aenet}.

\begin{table*}[!h]
	
	\centering \footnotesize
	\renewcommand \arraystretch{1.1}
	\caption{Parameter settings of evaluated methods \cite{fang2019nonlinear, fang2020aenet}}
	\label{table112}
	\begin{tabular}[b]{p{0.05cm}p{2.5cm}p{3.5cm}p{0.6cm}p{2cm}p{1cm}p{1cm}p{1.7cm}p{1.5cm}}
			\hline \\
		No.& Method &Parameters &Year  & Category  &Time(s)&Evolution&Robustness and Generality &Model Size(M) \\ \hline 
		1&FZL \cite{Lahoud2019FastZERO}&$r_b = 35, \varepsilon_b = 0.01, r_d =7$&2019&Hybrid&0.560&$\times$&$\checkmark$&$\times$ \\ 
		2&CSR \cite{Liu2016ImageCSR}&$\lambda=0.01$ &2016&Multi-scale& 47.199&$\times$&$\times$&$\times$ \\ 
		3&DL \cite{Li_2018DL}& $\alpha_1=\alpha_2=0.5, k\subset[1,2]$&2018& Deep learning &3.331&$\times$&$\times$&$549.00$\\ 
		4& DENSE \cite{Li2018DenseFuse}&$Epoach=4, Lr=0.0001$ &2019& Deep learing& 0.159&$\times$&$\times$&$1.02$\\ 
		5&FusionGAN \cite{Ma2018Infrared}&$Epoach=10, Lr=0.0001$ &2019&GAN&0.094&$\times$&$\times$&$7.24$\\ 
		6&IFCNN \cite{ZHANG202099}&$L_{r0}=0.01, power=0.9$&2020&Deep learning & {0.021}&$\times$&$\checkmark$ &$0.32(170)$\\
		7&DTCWT \cite{Lewis2007Pixel}&$\times$&2007& Wavelets & 0.148&$\times$&$\times$&$\times$\\ 
		8&LATLRR \cite{Li2018InfraredLTLRR}&$\lambda =0.4, stride=1$ &2020&Multi-scale  &30.897&$\times$&$\times$&$\times$\\ 
		9&LP-SR \cite{Liu2015ALPSR}&$overlap=6, \epsilon =0.1, level=4$&2015&Hybrid& 0.017&$\times$&$\times$&$\times$\\ 
		10&DSIFT \cite{efae}&$Scale=48, blocksize=8, matching=1$&2015&Other& 3.696 &$\times$&$\times$&$\times$\\ 
		11&CNN\_IV \cite{Liu2017InfraredCNN} & $t=0.6$ &2017&Hybrid& 21.594 &$\times$&$\times$&$1.54$\\ 
		12&CNN\_MF \cite{Liu2017MultiCNN} & $Momentum=0.9, decay=0.0005, threshold=0.5$ &2017&Hybrid& 50.225 &$\times$&$\times$&$30.90$\\ 	
		13&CVT \cite{Nencini2007RemoteCVT} &$is_real=1, finest=1$&2007&Multi-scale & 0.913&$\times$&$\times$&$\times$\\ 
		14&CBF \cite{Shreyamsha2015ImageCBF} &$\sigma s=1.8, \sigma r=25, ksize=11$&2015&Multi-scale&8.177&$\times$&$\times$&$\times$\\ 
		15&JSR \cite{Zhang2013Dictionary}&$Unit=7, step=1, dic_size=256, k=16$	&2013&Sparse representation& 95.670&$\times$&$\times$&$\times$\\ 
		16&JSRSD \cite{Liu2017InfraredJSR-SD} &$Unit=7, step=1, dic_size=256, k=16$&2017&Saliency-based& 126.041 &$\times$&$\times$&$\times$\\ 
		17&GTF \cite{Ma2016InfraredGTF}&$Epsr=epsf=tol=1, loops=5$ &2016&Other & 1.824 &$\times$&$\times$&$\times$\\ 
		18&WLS \cite{Ma2017InfraredWLS}&$\sigma_s=2, \sigma_r=0.05, nLevel=4$&2017&Hybrid& 8.397&$\times$&$\times$&$\times$\\ 
		19&RP \cite{Toet1989ImageRP}&$\times$&1989&Pyramid&0.057 &$\times$&$\times$&$\times$\\ 
		20&MSVD \cite{Naidu2011Image}  &$\times$&2011&Multi-scale& 0.207&$\times$&$\times$&$\times$\\ 
		21&MGFF \cite{Durga2019Multi}&$R=9, \varepsilon=10^3, k=4$&2019&Multi-scale  & 1.080&$\times$&$\times$&$\times$\\
		22 &ZCA \cite{Li2018Infrared}& $K=2, i=4 and i=5$  &2019&Hybrid&0.997 &$\times$&$\times$&$\times$\\
		23&ADF \cite{Bavirisetti2016Fusion} &$w1=w2=0.5$&$2016$&Multi-scale &$\times$&$\times$&$\times$&$\times$\\
		24&FPDE \cite{Bavirisetti2017Multi} &$At=0.9, n=20, k=4, \delta t=0.9$&$2017$&Subspace & $\times$&$\times$&$\times$&$\times$\\
		25&IFEVIP \cite{Zhang2017Infrared} &$Nd = 512, Md = 32, Gs = 9, MaxRatio = 0.001, StdRatio = 0.8 $&$2017$&Other & $\times$&$\times$&$\times$&$\times$\\	
		
		26&DeepFuse \cite{PrabhakarK.Ram2017DADU}&$ Epoach=100, lr=0.0001$&$2017$&Other & 0.101&$\times$&$\times$&$0.33$\\

		27&PGMI \cite{PGMI} &$Epoach=15, lr=1e-4, c_dim=1, stride=14, scale=3$&$2020$&Deep learning&0.044 &$\times$&$\times$&$0.80$\\
		28&FusionDN \cite{xu2020aaai} &$Ps=64, lam=80000, num=40$&$2020$&Deep learning & 0.850&$\times$&$\checkmark$&$1638.40(527)$\\
		29&SAF \cite{fang2019crossmodale} &$Epoach=256, bs=5, lr=0.000001$&$2020$&Deep learning &0.070 &$\times$&$\times$&$192.00(527)$\\
		
		30&NestFuse-max \cite{li2020nestfuse} &$Nb_{filter}=[64,112,160,208,256]$&$2020$&Deep learning &$0.116$ &$\times$&$\times$&$20.80$\\
		
		31&U2Fusion \cite{U2Fusion} &$num=30, epoach=[3,2,2], lam=0$&$2020$&Deep learning &0.857&$\times$&$\checkmark$&1064.96\\
		
		32&AE-Net \cite{fang2020aenet} &Dynamic&$2020$&Deep learning &0.071&$\checkmark$&$\checkmark$&192(527)\\
		
		33&\textbf{AE-Netv2-M}  &\textbf{Lr=adaptive, epoach=110, batchsize=16}&$\textbf{2020}$&\textbf{Deep learning} &\textbf{0.0128}&$\checkmark$&$\checkmark$&\textbf{0.896}\\
		
		34&\textbf{AE-Netv2-GCB} &\textbf{Lr=adaptive, epoach=110, batchsize=16}&$\textbf{2020}$&\textbf{Deep learning} &\textbf{0.0098}&$\textbf{\checkmark}$&$\textbf{\checkmark}$&\textbf{0.069}\\
		
		35&\textbf{AE-Netv2-CNN} &\textbf{Lr=adaptive, epoach=110, batchsize=16}&$\textbf{2020}$&\textbf{Deep learning} &\textbf{0.010}&$\textbf{\checkmark}$&$\textbf{\checkmark}$&\textbf{0.490}\\

		36&\textbf{AE-Netv2-Separable} &\textbf{Lr=adaptive, epoach=110, batchsize=16}&$\textbf{2020}$&\textbf{Deep learning} &\textbf{0.0097}&$\textbf{\checkmark}$&$\textbf{\checkmark}$&\textbf{0.11}\\

		37&\textbf{AE-Netv2-Squeeze} &\textbf{Lr=adaptive, epoach=110, batchsize=16}&$\textbf{2020}$&\textbf{Deep learning} &\textbf{0.0098}&$\textbf{\checkmark}$&$\textbf{\checkmark}$&\textbf{0.25}\\

		38&\textbf{AE-Netv2-Inception} &\textbf{Lr=adaptive, epoach=110, batchsize=16}&$\textbf{2020}$&\textbf{Deep learning} &\textbf{0.0108}&$\textbf{\checkmark}$&$\textbf{\checkmark}$&\textbf{0.91}\\
		
		39&\textbf{AE-Netv2-Squeeze-GCB} &\textbf{Lr=adaptive, epoach=110, batchsize=16}&$\textbf{2020}$&\textbf{Deep learning} &\textbf{0.0098}&$\textbf{\checkmark}$&$\textbf{\checkmark}$&\textbf{0.08}\\
		\hline
	\end{tabular}
\end{table*}

\textit{3) Methods:} Compared with AE-Net, this paper adds several latest image fusion method \cite{U2Fusion,li2020nestfuse,fang2019crossmodale} published in 2020. Other comparison image fusion methods are the same as \cite{fang2020aenet}, specific algorithms can be referred to Table~\ref{table112} and papers\cite{fang2020aenet}.

\textit{4) Implementation details:} 
Before the experiment, we need to clarify the following questions.
\textit{a)} In all subsequent experiments, we converted all images into grayscale images for subsequent image fusion \cite{fang2020aenet}. \textit{b)} For the time in Table~\ref{table112}, some data will change compared with AE-Net, because we find that the amount of VIFB \cite{zhang2020vifb} data is too small and the time used to describe the image fusion algorithm seems unfair. Therefore, we will normalize the image size to 400$\times$400, 8-bit grayscale image, 300 pieces to test the average time. For deep learning methods, in order to prevent the influence of network initialization on the first image, we will count the time from the second image. \textit{c)} Our experimental platform is desktop 3.0 GHZ i5-8500, RTX2070, 32G memory.


\subsection{Comparative Experiments}
In this section, in order to verify the robustness and generality of AE-Netv2, we will carry out comparative experiments and visual display on multi-exposure dataset, multi-focus dataset, medical dataset, infrared and visible dataset, combined vision system dataset. 

\subsubsection{Cross-modal Image Fusion}
\begin{figure}[ht]
	
	
	\includegraphics[scale=1,width=0.45\textwidth]{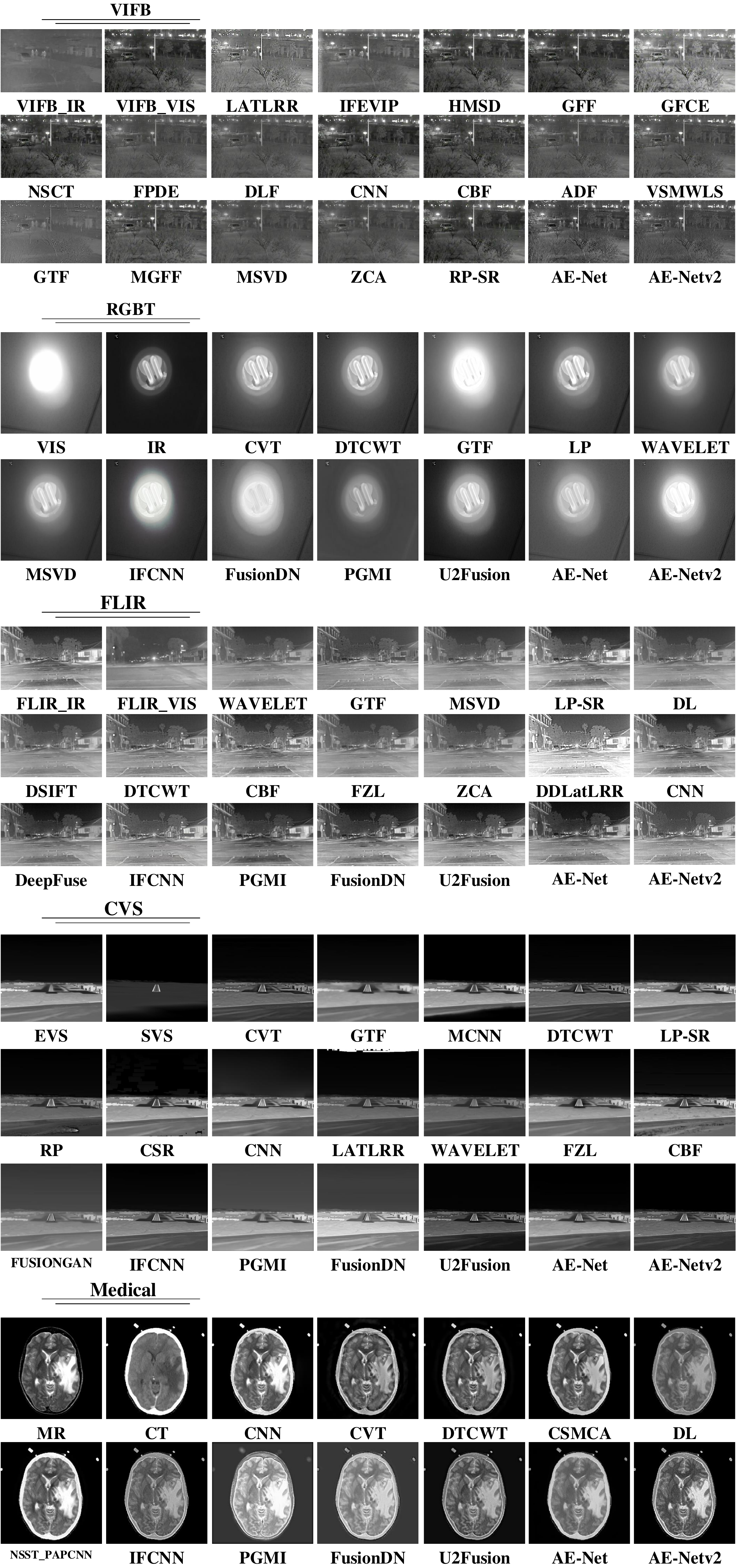}
	\centering
	
	\caption{
		Image fusion results of tested methods on six cross-modal datasets.}
	\label{f33}
\end{figure}
Our image fusion method is tested on cross-modal datasets which include FLIR, VIFB, RGBT1000 and CVS datasets. In the four data sets, except the fusion image of VIFB dataset is obtained directly, the other fusion images are obtained by testing according to the parameters of the original paper. In the experiment, AE-Net and AE-Netv2 have a very stable and robust fusion effect. From Fig.~\ref{f33}, we can find that AE-Netv2 image fusion method fully retains the detailed texture information of cross-modal images. \textbf{Encouragingly}, AE-Netv2 is better than the fusion effect of the latest published on AAAI \cite{PGMI, xu2020aaai} and PAMI \cite{U2Fusion} in 2020. We need to point out that the AE-Netv2 used here is the basic version of GCB and if other optimized architectures are used, the subjective image fusion effect will be better. See the analysis experiment specifically \ref{f7} \ref{f8}.

\subsubsection{Homomodality Image Fusion}
For homomodal image fusion tasks, we carry out experiments on multi-exposure and multi-focus image fusion datasets. In this experiment, except IFCNN, PGMI, FusionDN, U2Fusion and AE-Net, other fusion results are from MFIF \cite{ZhangXingchen2020MIFA} data benchmark.
\begin{figure}[ht]
	
	
	\includegraphics[scale=1,width=0.45\textwidth]{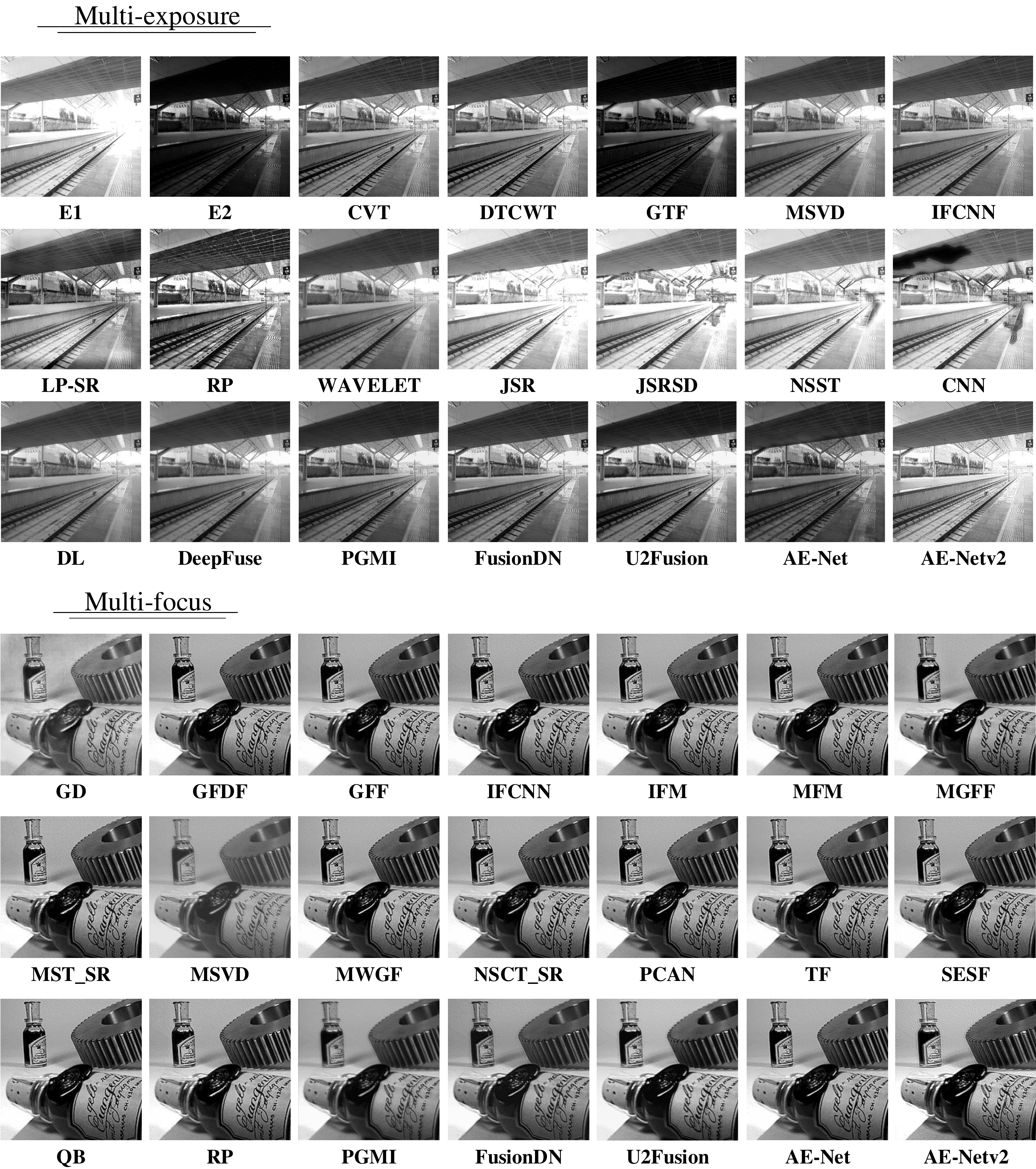}
	\centering
	
	\caption{
		Image fusion results of tested methods on multi-focus images.}
	\label{f4}
\end{figure}
From the experimental results, AE-Netv2 still has good result, although the results are not the best, but compared with IFCNN, FusionDN, PGMI, AE-Net and U2Fusion proposed in 2020, some noise is introduced. There are two reasons for this. On the one hand, there are errors in the relative optimal solution; on the other hand, multi-tasks are trained at the same time. Although the maximum collaborative optimization is achieved, the noise of different data distribution is also introduced to a certain extent. Of course, this is also closely related to the network architecture of conventional convolution kernel group convolution. 

\subsubsection{Objective Index Evaluation Comparative Experiment}
\begin{figure}[ht]
	
	
	\includegraphics[scale=1,width=0.45\textwidth]{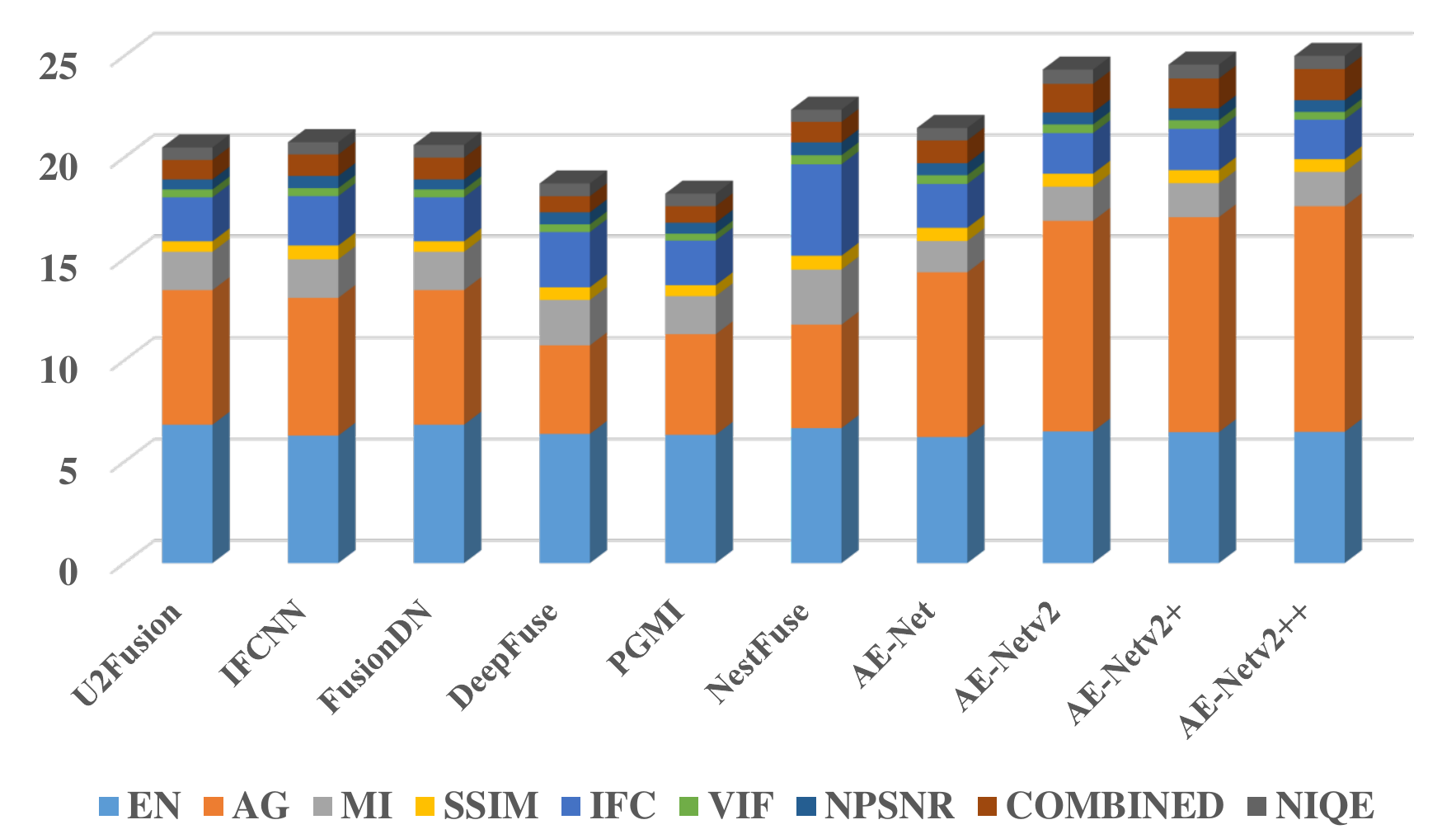}
	\centering
	
	\caption{Average objective index of five image fusion tasks.}
	\label{f55}
\end{figure}
In order to further demonstrate the superiority of AE-Netv2, we will use eight evaluation indexes to evaluate the latest deep learning image fusion methods. In this experiment, we use the mean value of 8 indexs of each image in cross-modal image fusion task, multi-focus image fusion task and multi-exposure image fusion tasks as comparison parameters. As there are many tested methods, we mainly compare the latest several image fusion methods, which the robustness and generality of these fusion algorithms have been greatly improved. In order to prevent the influence of PSNR index on other indexes, we normalized the PSNR index. In the same way, we also normalized the reciprocal of NIQE index. From Fig.~\ref{f55}, we can find that AE-Netv2 has better advantages than several latest deep learning image fusion methods. Even compared with AE-Net, the performance has been improved. In addition, we find that NestFuse method seems to have better metrics than AE-Net, FusionDN, U2Fusion et al. However, we observe the fused image, we find that NestFuse loses a lot of texture details and has the problem of local blur. This phenomenon shows that existing objective indicators can only reflect the image quality to a certain extent and fail to have a complete image quality characterization ability. We have also confirmed the correctness of this conclusion in previous related researchs~\cite{fang2019crossmodal}. In order to improve the image quality, we will show more subjective visual effects will be detailed in Fig.~\ref{f8}, including our proposed optimization network architectures and the algorithm published on AAAI and PAMI in 2020.

\subsubsection{Comparative Experiment of Multi-scale Methods Based on Deep Learning}
In this experiment, we will compare SAF, NestFuse and AE-Netv2\_M three image fusion methods on FLIR, Multi-focus, CVS, TNO public datasets. Therefore, these three methods directly or indirectly introduce multi-scale features.
\begin{figure}[ht]
	
	
	\includegraphics[scale=1,width=0.49\textwidth]{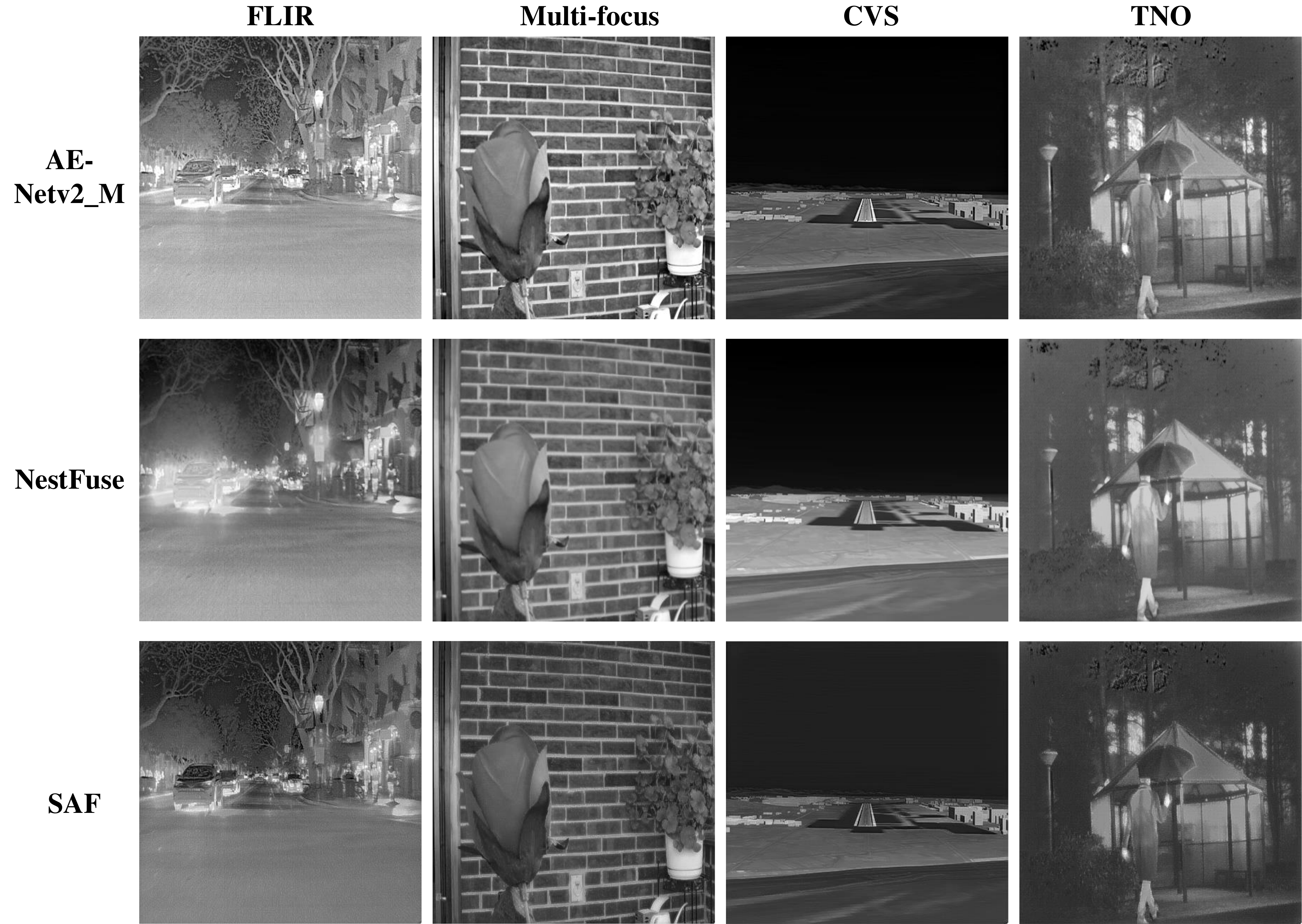}
	\centering
	
	\caption{Multi-scale image fusion methods based on deep learning.}
	\label{f99}
\end{figure}
From Fig.~\ref{f99}, we find that AE-Netv2\_M image fusion method with pooling layer has better image fusion effect than SAF~\cite{fang2019crossmodale} and NestFuse~\cite{li2020nestfuse}. Compared with all tested methods, our image fusion method has better performance on multiple datasets, which has rich texture details and good visual clarity. The experiment also fully shows that the introduction of pooling layer can still achieve good image fusion effect. Although NestFuse also introduces pooling layer, there is still a big gap between NestFuse and ours. The main reason for this gap is that NestFuse adopts the traditional similarity loss function while we adopt the brain like continuous learning optimization method. The former is to calculate the similarity while we are the real image quality optimization. Meanwhile, SAF method has better image quality than NestFuse and preserves more details of the image. This shows that SAF method has more advantages than NestFuse method with pool layers, but SAF has higher spatial complexity. 

\subsubsection{Comparative Analysis Experiment of Commonness and Characteristics}

\begin{figure}[ht]
	
	
	\includegraphics[scale=1,width=0.48\textwidth]{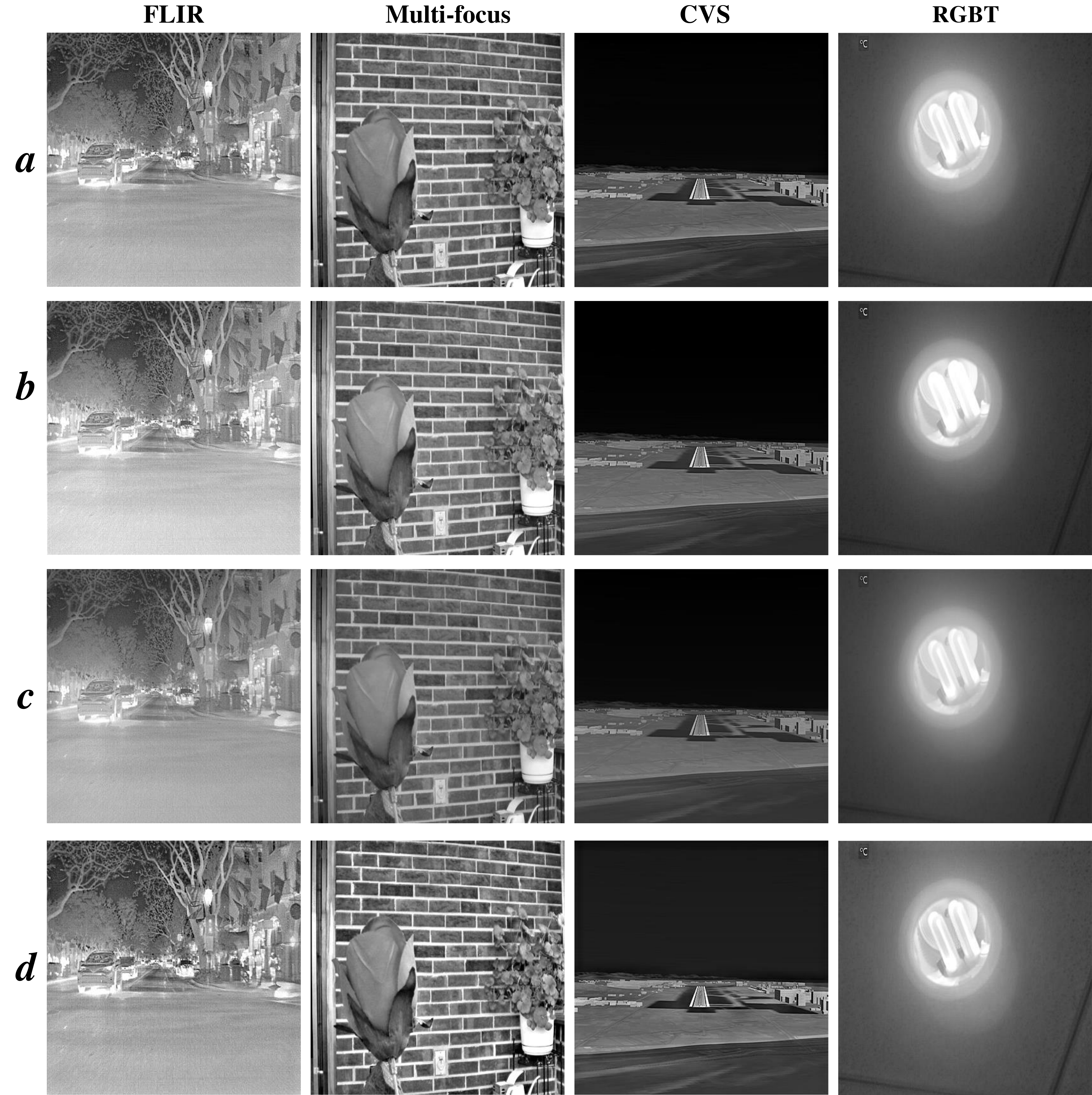}
	\centering
	
	\caption{Commonness and characteristics of different image fusion task, where a represents common image fusion result in hybrid training datsets; b represents image fusion result in CVS dataset; c represents image fusion result in multi-focus training dataset; d represents image fusion result in RGBT training dataset.}
	\label{f15}
\end{figure}
From Fig.~\ref{f15}, we find that the fusion results of common image fusion task have better subjective effect than that of single task fusion. Meanwhile, we find that this degradation is not obvious. The main reason is that we adopt the brain like continuous learning optimization method which can make different image fusion tasks have the same image quality definition instead of simple similarity measurement. In the experiment, we also find that if the traditional similarity loss is used to optimize the image fusion weight, the mixed data training will get worse image fusion effect. This is because this similarity index will increase the difference of data distribution of different image fusion tasks and reduce the commonness of image quality. Therefore, in practical experiments, we can find that most of the fusion results of traditional similarity loss training on a single dataset are better than that of hybrid training. Of course, the results of deep learning image fusion using traditional similarity loss are far from our results, whether on a single dataset or on a mixed dataset. The former focuses on similarity measurement, while AE-Netv2 focus on improving image quality.

\subsection{Analysis Experiments}
In addition, we also carry out analysis experiments for AE-Netv2. 
\textit{Firstly}, comparative analysis experiments using different network depths. \textit{Secondly}, comparative analysis of image fusion efficiency. \textit{Finally}, comparative analysis of different image fusion network architectures.

\begin{table*}[h]
	
	\centering \footnotesize
	\renewcommand \arraystretch{1.1}
	\caption{Network achitecture pameters}
	\label{table5}
	\begin{tabular}[b]{p{1.5cm}p{2cm}p{2cm}p{2cm}p{2cm}p{2cm}p{2cm}}
		\hline
		Type 	&IFCNN& U2Fusion& AE-Net&AE-Netv2  &AE-Netv2+ & AE-Netv2++  \\ \hline
		Flops		&20,797,920,000	  & 211,028,997,456 & 175,053,109,504&860,480,000  &6,584,640,000 & 245,920,000  \\	
		Parameters  & 83,587& 659,217 & 16,493,555&5,378  &41,154 & 1537  \\
		\hline
	\end{tabular}
\end{table*}
\begin{figure}[ht]
	
	
	\includegraphics[scale=1,width=0.48\textwidth]{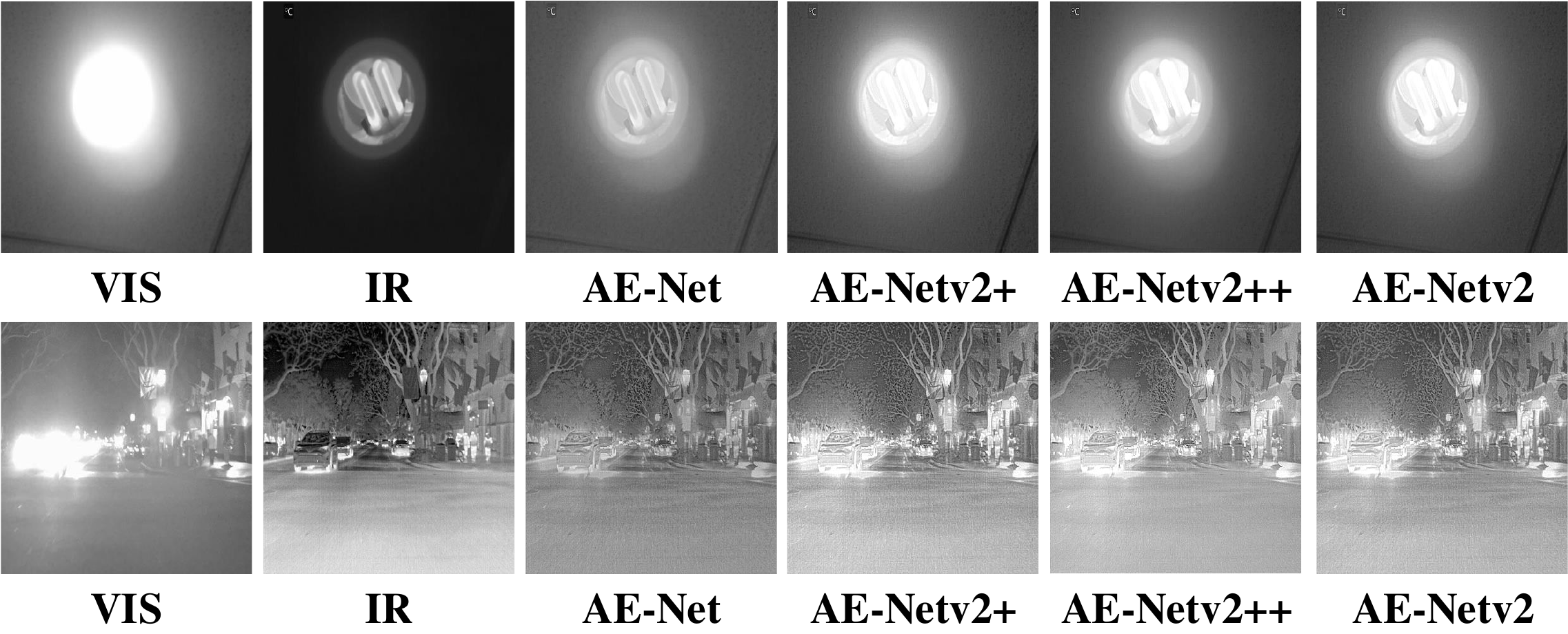}
	\centering
	
	\caption{Image fusion results of tested network architecture on RGBT and FLIR datasets.}
	\label{f6}
\end{figure}

\begin{table*}[h]
	\centering \footnotesize
	\renewcommand \arraystretch{1.1}
	\caption{Differnet network architecture}
	\label{table6}
	\begin{tabular}{c|c|cccccccc}
		\hline
		\multirow{8}{*}{Metrics} & Type     & Regular       & GCB           & Inception & Separable     & Squeeze & GCB+Inception & \multicolumn{1}{l}{Squeeze+GCB} & Squeeze2+GCB  \\ \cline{2-10} 
		& Combined & 1.35          & 1.29          & 1.33      & 1.42          & 1.61    & 1.26          & 1.61                             & \textbf{1.67} \\ \cline{2-2} 
		& NIQE     & \textbf{0.72} & 0.70          & 0.69      & 0.64          & 0.65    & 0.69          & 0.66                             & 0.63          \\ \cline{2-2} 
		& SSIM     & 0.63          & 0.62          & 0.63      & \textbf{0.64} & 0.63    & 0.63          & 0.63                             & 0.62          \\ \cline{2-2} 
		& EN       & 6.42          & 6.42          & 6.30      & 6.31          & 6.29    & 6.46          & 6.29                             & 6.33          \\ \cline{2-2} 
		& PSNR     & 0.48          & 0.48          & 0.48      & 0.48          & 0.48    & 0.48          & 0.48                             & 0.48          \\ \cline{2-2} 
		& BRENNER  & 5.63          & 5.20          & 5.55      & 6.15          & 7.53    & 5.00          & 7.48                             & \textbf{7.97} \\ \cline{2-2} 
		& VIFF     & 0.32          & 0.34          & 0.34      & 0.33          & 0.32    & \textbf{0.35} & 0.32                             & 0.31          \\ \hline
		\multicolumn{2}{c}{Time(ms)}      &10.17          & 9.83         & 10.83     & \textbf{9.67} & 9.83    & 11.67         & 9.83                             & 13.83          \\ \hline
		\multicolumn{2}{c}{Size(M)}       & 0.48          & \textbf{0.07} & 0.91      & 0.11          & 0.25    & 0.11          & 0.08                             & 0.10           \\ \hline
	\end{tabular}
\end{table*}
\subsubsection{Performance Analysis Experiment of Different Network Depths}
\begin{figure*}[h]
	
	
	\includegraphics[scale=0.1,width=1\textwidth]{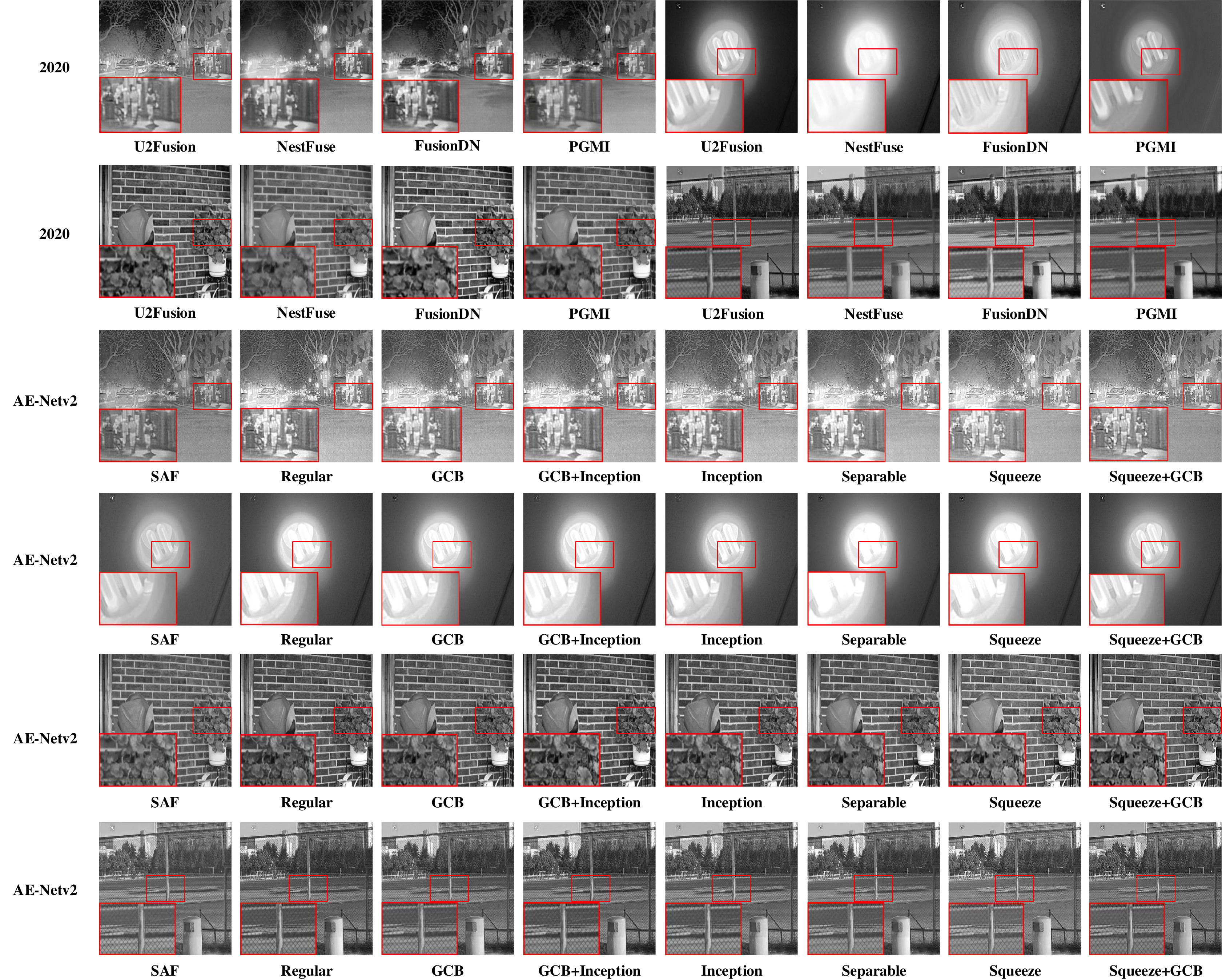}
	\centering
	
	\caption{Performance comparison of AE-Netv2 with different network architectures and several state-of-the-art methods in 2020.}
	\label{f8}
\end{figure*}

In this experiment, we mainly explore the influence of different depths on image fusion quality. We mainly carry out four architecture of comparative experiments, namely AE-Net based on SAF after prunning~\cite{fang2019crossmodale}, AE-Netv2, AE-Netv2+ and AE-Netv2++ in this paper. Compared with AE-Netv2, AE-Netv2+ replaces group convolution with regular convolution and other architecture modules is the same. AE-Netv2++ is to replace group convolution with regular convolution. In addition, remove hop connection. From Fig.~\ref{f6} and Fig.~\ref{f8}, we can see that several simplified architectures of AE-Netv2 can achieve better image fusion quality than tested methods. Experimental results show that the change of network depth has limited influence on image fusion quality and shallow neural network can still obtain very good image fusion quality.

\begin{figure}[h]
	
	
	\includegraphics[scale=1,width=0.45\textwidth]{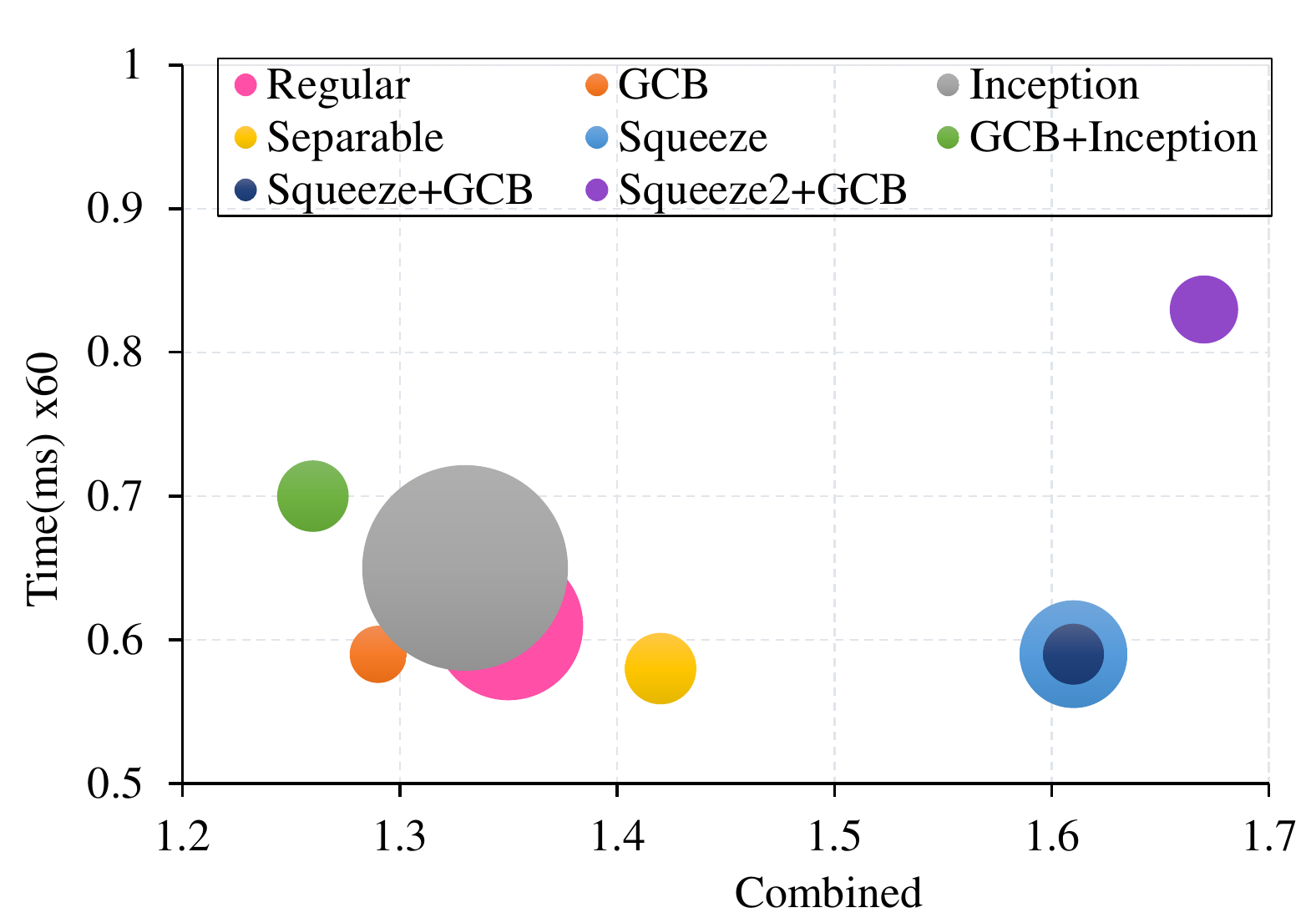}
	\centering
	
	\caption{Image fusion quality, efficiency and model size distribution of AE-Netv2 with different network architectures.}
	\label{f7}
\end{figure}
\subsubsection{Performance Analysis Experiments of Image Fusion Efficiency}
In this experiment, we will analyze the efficiency of different network architectures, including time complexity, spatial complexity and computational complexity. In this paper, we first analyze the efficiency of different image fusion methods and then analyze the efficiency of different network architectures of AE-Netv2. The time statistics of tested methods are shown in Table~\ref{table112}. The data in brackets is the model size of pre-training. The distribution of image quality, fusion time and model size is shown in Fig.~\ref{f7}. The horizontal axis represents the image quality, the vertical axis represents the time and the circle size represents the size of the model. The larger the radius, the larger the model size. Because there are many contrast algorithms, we only compare the latest image fusion methods in 2020 for the calculation complexity. The details of our proposed method are shown in Fig.~\ref{f7} and Table~\ref{table6}. From Table~\ref{table112} and Fig.~\ref{f7}, we can see that AE-Netv2 has an remarkable advantage in image fusion efficiency and image fusion effect of tested methods. Compared with the fastest algorithm, the speed is increased by 2.14 times. Compared with the smallest deep learning model, the size is reduced by 11.59 times. The model size reflects the parameters that can be trained to a certain extent. The larger the parameters, the larger the model size will be. As shown in \ref{table5}, compared with the latest deep learning method, AE-Netv2 has less parameters and lower computational complexity. 

\subsubsection{Performance Analysis Experiment of Different Network Architectures}
In the first analytical experiment, we explore the influence of group convolution and different convolution depths on AE-Netv2. We will explore the influence of different network architectures on image fusion quality, mainly analyze and verify the regular convolution, group convolution, separable convolution, squeeze module, inception module and combined module. In this experiment, we will select the mean value of the objective indicators of five image fusion tasks as the evaluation criteria. In the same way, NIQE and PSNR are normalized and NIQE is reversed. From Table~\ref{table6}, we can find that the combination of CNN and GCB alone has great advantages in time and space, but there are still some gaps in performance compared with other network architectures. Compared with the method of combining compressed convolution with group convolution, the model size is greatly reduced when the fusion effect and time are the same. With the increase of the number of squeeze convolutions, its performance is slightly improved, but its time and space complexity is greatly improved. Therefore, considering the performance of each other's structure, the network architectures of squeeze convolution and group convolution is the best choice considering time, space and fusion quality. As shown in Fig.~\ref{f8}, we can find that AE-Netv2 can achieve remarkable results in terms of performance, time and space complexity by using squeeze convolution and group convolution. AE-Netv2 not only has good objective evaluation results, but also preserves the texture details of subjective effect very well. Compared with several state-of-the-art methods proposed in 2020, AE-Netv2 has better fusion quality. The change of image quality caused by the optimization strategy of AE-Netv2 optimal solution is obvious, which is not available in all tested methods. Therefore, a good optimization strategy can make the very simple network architecture product state-of-the-art result. This is also why the human brain can be so efficient and robust to complete a variety of visual processing tasks.






\section{Discussion}
\label{dis}
Comprehensive experiments in Sect.~\ref{setup} verify that small sample data and simple network architecture can also achieve the high efficiency, robustness and generality of image fusion. The reasons are as follows:

\textit{1) Autonomous Evolution Image Fusion Method}. This method is different from neural architecture search (NAS)~\cite{Nas} which is adopted by autonomous evolution. Compared with NAS, it has better continuous autonomous evolution ability and faster predict speed. It uses the complementary characteristics of existing methods as the prior knowledge of each iteration to guide network learning. Existing image fusion methods whether traditional methods or image fusion methods based on deep learning have their inherent advantages and disadvantages. Therefore, how to make full use of their advantages to form complementary advantages plays an important role in improving the robustness and generality of image fusion. From a new point of view, we propose to use a variety of image fusion algorithms to build an algorithm library, through which the optimal image fusion solution can be obtained. By adding different new algorithms, the performance of the network can be continuously improved and the network has the ability of continuous learning. This is the lack of existing image fusion theory.

\textit{2) Architecture Design of Image Fusion Network}. Existing image fusion methods have paid few research attention to image fusion efficiency. By contrast, in order to improve image fusion quality, researchers make the network have very high complexity. This is disadvantageous to the development of image fusion technology. Even though some image fusion methods only use some pre-training model weights, the size of the pre-training model seriously limits the development of image fusion technology. Therefore, we focus on the relationship of network architecture, efficiency and fusion quality of image fusion. By balancing the relationship among the three, we construct a very efficient, robust and universal image fusion architecture.

Why did the exploration of commonness and characteristics not reach the state of the art? Is it meaningless?

\textit{1) It is difficult to model loss function.} In this case, different image fusion tasks lack common description of image quality, which makes different data distribution unable to be unified. After feature transfer, feature forgetting will occur. \textit{2) Complementary characteristics of methods.} Because our method adopts AE-Netv2 architecture, the image fusion quality can be improved continuously through the complementary of different image fusion methods. \textit{3) Due to the superiority of AE-Netv2 architecture, the extracted common features are better than single task}. By comparing the subjective fusion effect with the most advanced image fusion algorithm in 2020, we can also find the correctness of this viewpoint. 

It is very meaningful for us to explore the commonness and characteristics of different image fusion tasks, which provides direction and reference for the future development of image fusion field.


\section{Conclusions and Future Work}
\label{con}
Inspired by human brain cognitive mechanism, we proposed an efficient image fusion method with the ability of continuous learning. \textit{The main differences between our image fusion method and existing image fusion methods are as follows \cite{fang2020aenet}}. \textbf{Firstly}, we explored the hidden relationship of image fusion in network architecture, time efficiency and fusion quality for the first time. \textbf{Secondly}, inspired by other visual tasks, we explored the influence of pooling layer on image fusion task. \textbf{Finally}, we explored the commonness and characteristics of different image fusion tasks, which provides a research basis for this field. 

In order to verify the impact of different network architectures on fusion quality and fusion efficiency, we conducted exhaustive experiments on different network architectures. The research we have done demonstrates that AE-Netv2 is superior to other state-of-the-art methods in efficiency, robustness and generality. In addition, efficient and robust image fusion can be achieved through small sample data and simple network architecture. The pooling layer can also achieve good results when applied to image fusion tasks. Of course, our image fusion method also fully proves that there are some problems in the cross-modal image fusion based on the traditional similarity loss function or evaluating the cross-modal image quality by similarity. The image fusion quality is different from the similarity measurement. What we should explore is to improve the image quality, not the similarity with the original image. Our future work will continue to focus on image fusion quality and human brain characteristics.

\section*{Acknowledgment}
We appreciate the support of relevant funds.

\ifCLASSOPTIONcaptionsoff
  \newpage
\fi

\bibliographystyle{IEEEtran}
\bibliography{mybibfile}

\begin{thebibliography}{10}
\providecommand{\url}[1]{#1}
\csname url@samestyle\endcsname
\providecommand{\newblock}{\relax}
\providecommand{\bibinfo}[2]{#2}
\providecommand{\BIBentrySTDinterwordspacing}{\spaceskip=0pt\relax}
\providecommand{\BIBentryALTinterwordstretchfactor}{4}
\providecommand{\BIBentryALTinterwordspacing}{\spaceskip=\fontdimen2\font plus
\BIBentryALTinterwordstretchfactor\fontdimen3\font minus
  \fontdimen4\font\relax}
\providecommand{\BIBforeignlanguage}[2]{{%
\expandafter\ifx\csname l@#1\endcsname\relax
\typeout{** WARNING: IEEEtran.bst: No hyphenation pattern has been}%
\typeout{** loaded for the language `#1'. Using the pattern for}%
\typeout{** the default language instead.}%
\else
\language=\csname l@#1\endcsname
\fi
#2}}
\providecommand{\BIBdecl}{\relax}
\BIBdecl

\bibitem{fang2020aenet}
A.~Fang, X.~Zhao, J.~Yang, S.~Cao, and Y.~Zhang, ``Ae-net: Autonomous evolution
  image fusion method inspired by human cognitive mechanism,'' 2020.

\bibitem{MillerCPortex}
E.~K. Miller and J.~D. Cohen, ``An integrative theory of prefrontal cortex
  function,'' \emph{Annual Review of Neuroscience}, 2001.

\bibitem{GuangYang2009Smds}
G.~Yang, F.~Pan, and W.-B. Gan, ``Stably maintained dendritic spines are
  associated with lifelong memories,'' \emph{Nature}, 2009.

\bibitem{fang2019crossmodal}
A.~Fang, X.~Zhao, J.~Yang, and Y.~Zhang, ``\BIBforeignlanguage{eng}{A
  cross-modal image fusion method guided by human visual characteristics},''
  \emph{\BIBforeignlanguage{eng}{arXiv.org}}, 2020.

\bibitem{KochC1985Sisv}
C.~Koch and S.~Ullman, ``Shifts in selective visual attention: towards the
  underlying neural circuitry,'' \emph{Human neurobiology}, 1985.

\bibitem{Valois1996Visual}
D.~Valois and K.~K., ``Visual perception: Foundations of vision.''
  \emph{Science}, vol. 271, no. 5254, pp. 1371a--1371a, 1996.

\bibitem{Treisman1980A}
A.~M. Treisman and G.~Gelade, ``A feature-integration theory of attention,''
  \emph{Cognitive Psychology}, vol.~12, no.~1, pp. 97--136, 1980.

\bibitem{Arriaga2015Visual}
R.~I. Arriaga, D.~Rutter, M.~Cakmak, and S.~S. Vempala, ``Visual categorization
  with random projection,'' \emph{Neural Computation}, vol.~27, no.~10, pp.
  2132--2147, 2015.

\bibitem{Liu2017InfraredCNN}
Y.~Liu, X.~Chen, J.~Cheng, H.~Peng, and Z.~Wang, ``Infrared and visible image
  fusion with convolutional neural networks,'' \emph{International Journal of
  Wavelets Multiresolution \& Information Processing}, vol.~16, no.~3, pp.
  1--20, 2017.

\bibitem{Liu2017MultiCNN}
Y.~Liu, X.~Chen, H.~Peng, and Z.~Wang, ``Multi-focus image fusion with a deep
  convolutional neural network,'' \emph{Information Fusion}, vol.~36, pp.
  191--207, 2017.

\bibitem{Yu2017medicalCNN}
L.~Yu, C.~Xun, J.~Cheng, and P.~Hu, ``A medical image fusion method based on
  convolutional neural networks,'' in \emph{International Conference on
  Information Fusion}, 2017.

\bibitem{Yin2018MedicalNSSPAPCNN}
M.~Yin, L.~Xiaoning, L.~Yu, and C.~Xun, ``Medical image fusion with
  parameter-adaptive pulse coupled-neural network in nonsubsampled shearlet
  transform domain,'' \emph{IEEE Transactions on Instrumentation \&
  Measurement}, vol.~68, no.~1, pp. 1--16, 2018.

\bibitem{DengXin2020DCNN}
X.~Deng and P.~L. Dragotti, ``\BIBforeignlanguage{eng}{Deep convolutional
  neural network for multi-modal image restoration and fusion},''
  \emph{\BIBforeignlanguage{eng}{IEEE Transactions on Pattern Analysis and
  Machine Intelligence}}, pp. 1--1, 2020.

\bibitem{xu2020aaai}
H.~Xu, J.~Ma, Z.~Le, J.~Jiang, and X.~Guo, ``Fusiondn: A unified densely
  connected network for image fusion,'' in \emph{2020 34th AAAI Conference on
  Artificial Intelligence}, 2020.

\bibitem{PGMI}
Y.~X. X.~G. Hao~Zhang, Han~Xu and J.~Ma, ``Rethinking the image fusion: A fast
  unified image fusion network based on proportional maintenance of gradient
  and intensity.''\hskip 1em plus 0.5em minus 0.4em\relax IEEE, Feb 2020.

\bibitem{Li2018Infrared}
H.~Li, X.~jun Wu, and T.~S. Durrani, ``Infrared and visible image fusion with
  resnet and zero-phase component analysis,'' \emph{Infrared Physics \$
  Technology}, vol. 102, p. 103039, 2019.

\bibitem{MaFusionGAN}
M.~Jiayi, Y.~Wei, L.~Pengwei, L.~Chang, and J.~Junjun, ``Fusiongan: A
  generative adversarial network for infrared and visible image fusion,''
  \emph{Information Fusion}, vol.~48, pp. 11 -- 26, 2019.

\bibitem{fang2019crossmodale}
A.~Fang, X.~Zhao, and Y.~Zhang, ``\BIBforeignlanguage{eng}{Cross-modal image
  fusion guided by subjective visual attention},''
  \emph{\BIBforeignlanguage{eng}{Neurocomputing}}, 2020.

\bibitem{ZHANG202099}
Y.~Zhang, Y.~Liu, P.~Sun, H.~Yan, X.~Zhao, and L.~Zhang, ``Ifcnn: A general
  image fusion framework based on convolutional neural network,''
  \emph{Information Fusion}, vol.~54, pp. 99 -- 118, 2020.

\bibitem{li2020nestfuse}
H.~Li, X.-J. Wu, and T.~Durrani, ``Nestfuse: An infrared and visible image
  fusion architecture based on nest connection and spatial/channel attention
  models,'' \emph{IEEE Transactions on Instrumentation and Measurement}, 2020.

\bibitem{U2Fusion}
J.~J. X.~G. Han~Xu, Jiayi~Ma* and H.~Ling, ``U2fusion: A unified unsupervised
  image fusion network,'' \emph{IEEE Transactions on Pattern Analysis and
  Machine Intelligence}, 2020.

\bibitem{fang2019nonlinear}
A.~Fang, X.~Zhao, J.~Yang, and Y.~Zhang, ``Non-linear and selective fusion of
  cross-modal images,'' 2019.

\bibitem{Zhao2017Multisensor}
W.~Zhao, H.~Lu, and W.~Dong, ``Multisensor image fusion and enhancement in
  spectral total variation domain,'' \emph{IEEE Transactions on Multimedia},
  vol.~PP, no.~99, pp. 1--1, 2017.

\bibitem{Cui2015Detail}
G.~Cui, H.~Feng, Z.~Xu, Q.~Li, and Y.~Chen, ``Detail preserved fusion of
  visible and infrared images using regional saliency extraction and
  multi-scale image decomposition,'' \emph{Optics Communications}, vol. 341,
  no. 341, pp. 199--209, 2015.

\bibitem{ZhangInfrared}
Z.~Xiaoye, M.~Yong, F.~Fan, Z.~Ying, and H.~Jun, ``Infrared and visible image
  fusion via saliency analysis and local edge-preserving multi-scale
  decomposition,'' \emph{Journal of the Optical Society of America. A, Optics,
  Image Science, and Vision}, vol.~34, no.~8, pp. 1400--1410, 2017.

\bibitem{Bhatnagar}
G.~Bhatnagar, Q.~M.~J. Wu, and Z.~Liu, ``\BIBforeignlanguage{eng}{Directive
  contrast based multimodal medical image fusion in nsct domain},''
  \emph{\BIBforeignlanguage{eng}{IEEE Transactions on Multimedia}}, vol.~15,
  no.~5, pp. 1014--1024, 2013.

\bibitem{Bavirisetti2016Two}
D.~P. Bavirisetti and R.~Dhuli, ``Two-scale image fusion of visible and
  infrared images using saliency detection,'' \emph{Infrared Physics \&
  Technology}, vol.~76, pp. 52--64, 2016.

\bibitem{Zhang2015A}
B.~Zhang, X.~Lu, H.~Pei, and Y.~Zhao, ``A fusion algorithm for infrared and
  visible images based on saliency analysis and non-subsampled shearlet
  transform,'' \emph{Infrared Physics \& Technology}, vol.~73, pp. 286--297,
  2015.

\bibitem{Liu2017InfraredJSR-SD}
C.~Liu, Y.~Qi, and W.~Ding, ``Infrared and visible image fusion method based on
  saliency detection in sparse domain,'' \emph{Infrared Physics \& Technology},
  vol.~83, pp. 94--102, 2017.

\bibitem{Ma2017InfraredWLS}
J.~Ma, Z.~Zhou, B.~Wang, and H.~Zong, ``Infrared and visible image fusion based
  on visual saliency map and weighted least square optimization,''
  \emph{Infrared Physics \& Technology}, vol.~82, pp. 8--17, 2017.

\bibitem{LIU202071}
Y.~Liu, L.~Wang, J.~Cheng, C.~Li, and X.~Chen, ``Multi-focus image fusion: A
  survey of the state of the art,'' \emph{Information Fusion}, vol.~64, pp. 71
  -- 91, 2020.

\bibitem{PrabhakarK.Ram2017DADU}
K.~R. Prabhakar, V.~S. Srikar, and R.~V. Babu, ``\BIBforeignlanguage{eng}{A
  deep unsupervised approach for exposure fusion with extreme exposure image
  pairs},'' in \emph{\BIBforeignlanguage{eng}{2017 IEEE International
  Conference on Computer Vision}}.\hskip 1em plus 0.5em minus 0.4em\relax IEEE,
  2017.

\bibitem{MPCNN}
M.~Y.~D. Wang Z~B, ``Medical image fusion using m-pcnn,'' \emph{Information
  Fusion}, vol. 9(2), pp. 176--185, 2008.

\bibitem{GiuseppePansharpening}
M.~Giuseppe, C.~Davide, V.~Luisa, and S.~Giuseppe, ``Pansharpening by
  convolutional neural networks,'' \emph{Remote Sensing}, vol.~8, no.~7, p.
  594.

\bibitem{MA202085}
J.~Ma, P.~Liang, W.~Yu, C.~Chen, and J.~Jiang, ``Infrared and visible image
  fusion via detail preserving adversarial learning,'' \emph{Information
  Fusion}, vol.~54, pp. 85--98, 2019.

\bibitem{Li2018DenseFuse}
H.~Li and X.-J. Wu, ``\BIBforeignlanguage{eng}{Densefuse: A fusion approach to
  infrared and visible images},'' \emph{\BIBforeignlanguage{eng}{IEEE
  Transactions on Image Processing}}, vol.~28, no.~5, pp. 2614--2623, 2019.

\bibitem{PANGAN}
C.~C. P. L. e.~a. Jiayi~Ma, Wei~Yu, ``Pan-gan: An unsupervised pan-sharpening
  method for remote sensing image fusion,'' \emph{Information Fusion}, vol.~62,
  no.~1, pp. 110--120, 20202.

\bibitem{IandolaForrest2016SAaw}
F.~Iandola, H.~Song, and M.~e.~a. Moskewicz,
  ``\BIBforeignlanguage{eng}{Squeezenet: Alexnet-level accuracy with 50x fewer
  parameters and <0.5mb model size},''
  \emph{\BIBforeignlanguage{eng}{arXiv.org}}, 2016.

\bibitem{Han2016Deep}
S.~Han, H.~Mao, and W.~J. Dally, ``Deep compression: Compressing deep neural
  networks with pruning, trained quantization and huffman coding,'' in
  \emph{ICLR}, 2016.

\bibitem{HintonGeoffrey2015DtKi}
G.~Hinton, O.~Vinyals, and J.~Dean, ``\BIBforeignlanguage{eng}{Distilling the
  knowledge in a neural network},'' \emph{\BIBforeignlanguage{eng}{arXiv.org}},
  2015.

\bibitem{RastegariMohammad2016XICU}
M.~Rastegari, V.~Ordonez, J.~Redmon, and A.~Farhadi, ``Xnor-net: Imagenet
  classification using binary convolutional neural networks,'' 2016.

\bibitem{ZhangXiangyu2017SAEE}
X.~Zhang, X.~Zhou, M.~Lin, and J.~Sun, ``\BIBforeignlanguage{eng}{Shufflenet:
  An extremely efficient convolutional neural network for mobile devices},''
  \emph{\BIBforeignlanguage{eng}{arXiv.org}}, 2017.

\bibitem{1576816}
H.~R. Sheikh and A.~C. Bovik, ``Image information and visual quality,''
  \emph{IEEE Transactions on Image Processing}, vol.~15, no.~2, pp. 430--444,
  Feb 2006.

\bibitem{1284395}
Z.~Wang, A.~C. Bovik, H.~R. Sheikh, and E.~P. Simoncelli, ``Image quality
  assessment: from error visibility to structural similarity,'' \emph{IEEE
  Transactions on Image Processing}, vol.~13, no.~4, pp. 600--612, April 2004.

\bibitem{Han2013A}
Y.~Han, Y.~Cai, Y.~Cao, and X.~Xu, ``A new image fusion performance metric
  based on visual information fidelity,'' \emph{Information Fusion}, vol.~14,
  no.~2, pp. 127--135, 2013.

\bibitem{MittalA2013MaCB}
A.~Mittal, R.~Soundararajan, and A.~C. Bovik, ``\BIBforeignlanguage{eng}{Making
  a completely blind image quality analyzer},''
  \emph{\BIBforeignlanguage{eng}{IEEE Signal Processing Letters}}, vol.~20,
  no.~3, pp. 209--212, 2013.

\bibitem{SijbersJ1996Qaio}
J.~Sijbers, P.~Scheunders, N.~Bonnet, D.~Van~Dyck, and E.~Raman,
  ``\BIBforeignlanguage{eng}{Quantification and improvement of the
  signal-to-noise ratio in a magnetic resonance image acquisition procedure},''
  \emph{\BIBforeignlanguage{eng}{Magnetic Resonance Imaging}}, vol.~14, no.~10,
  pp. 1157--1163, 1996.

\bibitem{Qu2002Information}
G.~Qu, D.~Zhang, and P.~Yan, ``Information measure for performance of image
  fusion,'' \emph{Electronics Letters}, vol.~38, no.~7, pp. 313--315, 2002.

\bibitem{JianruiCai2018LaDS}
J.~Cai, S.~Gu, and L.~Zhang, ``\BIBforeignlanguage{eng}{Learning a deep single
  image contrast enhancer from multi-exposure images},''
  \emph{\BIBforeignlanguage{eng}{IEEE Transactions on Image Processing}},
  vol.~27, no.~4, pp. 2049--2062, 2018.

\bibitem{Nejati2015Multi}
M.~Nejati, S.~Samavi, and S.~Shirani, ``Multi-focus image fusion using
  dictionary-based sparse representation,'' \emph{Information Fusion}, vol.~25,
  pp. 72--84, 2015.

\bibitem{ZhangXingchen2020MIFA}
X.~Zhang, ``Multi-focus image fusion: A benchmark,'' 2020.

\bibitem{Summers2003Harvard}
Summers and D, ``Harvard whole brain atlas,'' \emph{Journal of Neurology
  Neurosurgery \& Psychiatry}, vol.~74, no.~3, pp. 288--288, 2003.

\bibitem{zhang2020vifb}
X.~Zhang, P.~Ye, and G.~Xiao, ``Vifb: A visible and infrared image fusion
  benchmark,'' 2020.

\bibitem{FLIR}
AZoSensors, ``Flir releases starter thermal imaging dataset for machine
  learning advanced driver assistance development,'' 2018.

\bibitem{LiChenglong2019RotB}
C.~Li, X.~Liang, Y.~Lu, N.~Zhao, and J.~Tang, ``\BIBforeignlanguage{eng}{Rgb-t
  object tracking: Benchmark and baseline},''
  \emph{\BIBforeignlanguage{eng}{Pattern Recognition}}, vol.~96, 2019.

\bibitem{Lahoud2019FastZERO}
F.~Lahoud and S.~Susstrunk, ``\BIBforeignlanguage{eng}{Zero-learning fast
  medical image fusion},'' in \emph{\BIBforeignlanguage{eng}{2019 22th
  International Conference on Information Fusion (FUSION)}}.\hskip 1em plus
  0.5em minus 0.4em\relax ISIF - International Society of Information Fusion,
  2019, pp. 1--8.

\bibitem{Liu2016ImageCSR}
Y.~Liu, X.~Chen, R.~Ward, and Z.~J. Wang, ``Image fusion with convolutional
  sparse representation,'' \emph{IEEE Signal Processing Letters}, vol.~23,
  no.~12, pp. 1882--1886, 2016.

\bibitem{Li_2018DL}
H.~Li, X.-J. Wu, and J.~Kittler, ``Infrared and visible image fusion using a
  deep learning framework,'' \emph{2018 24th International Conference on
  Pattern Recognition}, Aug 2018.

\bibitem{Ma2018Infrared}
J.~Ma, Y.~Ma, and C.~Li, ``Infrared and visible image fusion methods and
  applications: A survey,'' \emph{Information Fusion}, vol.~45, pp. 153 -- 178,
  2019.

\bibitem{Lewis2007Pixel}
J.~J. Lewis, R.~J. O’Callaghan, S.~G. Nikolov, D.~R. Bull, and
  N.~Canagarajah, ``Pixel- and region-based image fusion with complex
  wavelets,'' \emph{Information Fusion}, vol.~8, no.~2, pp. 119--130, 2007.

\bibitem{Li2018InfraredLTLRR}
H.~Li, X.-J. Wu, and J.~Kittler, ``\BIBforeignlanguage{eng}{Mdlatlrr: A novel
  decomposition method for infrared and visible image fusion},''
  \emph{\BIBforeignlanguage{eng}{IEEE Transactions on Image Processing}},
  vol.~29, pp. 4733--4746, 2020.

\bibitem{Liu2015ALPSR}
Y.~Liu, S.~Liu, and Z.~Wang, ``A general framework for image fusion based on
  multi-scale transform and sparse representation,'' \emph{Information Fusion},
  vol.~24, pp. 147--164, 2015.

\bibitem{efae}
Y.~Liu, S.~Liu, and W.~et~al., ``\BIBforeignlanguage{eng}{Multi-focus image
  fusion with dense sift},'' \emph{\BIBforeignlanguage{eng}{Information
  Fusion}}, vol.~23, no.~C, pp. 139--155, 2015.

\bibitem{Nencini2007RemoteCVT}
F.~Nencini, A.~Garzelli, S.~Baronti, and L.~Alparone, ``Remote sensing image
  fusion using the curvelet transform,'' \emph{Information Fusion}, vol.~8,
  no.~2, pp. 143--156, 2007.

\bibitem{Shreyamsha2015ImageCBF}
S.~Kumar and B.~K., ``Image fusion based on pixel significance using cross
  bilateral filter,'' \emph{Signal Image \& Video Processing}, vol.~9, no.~5,
  pp. 1193--1204, 2015.

\bibitem{Zhang2013Dictionary}
Q.~Zhang, Y.~Fu, H.~Li, and J.~Zou, ``Dictionary learning method for joint
  sparse representation-based image fusion,'' \emph{Optical Engineering},
  vol.~52, no.~5, p. 7006, 2013.

\bibitem{Ma2016InfraredGTF}
J.~Ma, C.~Chen, C.~Li, and J.~Huang, ``Infrared and visible image fusion via
  gradient transfer and total variation minimization,'' \emph{Information
  Fusion}, vol.~31, no.~C, pp. 100--109, 2016.

\bibitem{Toet1989ImageRP}
A.~Toet, ``Image fusion by a ratio of low-pass pyramid,'' \emph{Pattern
  Recognition Letters}, vol.~9, no.~4, pp. 245--253, 1989.

\bibitem{Naidu2011Image}
V.~P.~S. Naidu, ``Image fusion technique using multi-resolution singular value
  decomposition,'' \emph{Defence Science Journal}, vol.~61, no.~5, pp.
  479--484, 2011.

\bibitem{Durga2019Multi}
D.~Bavirisetti, G.~Xiao, J.~Zhao, R.~Dhuli, and G.~Liu,
  ``\BIBforeignlanguage{eng}{Multi-scale guided image and video fusion: A fast
  and efficient approach},'' \emph{\BIBforeignlanguage{eng}{Circuits, Systems,
  and Signal Processing}}, vol.~38, no.~12, pp. 5576--5605, 2019.

\bibitem{Bavirisetti2016Fusion}
D.~P. Bavirisetti and R.~Dhuli, ``\BIBforeignlanguage{eng}{Fusion of infrared
  and visible sensor images based on anisotropic diffusion and karhunen-loeve
  transform},'' \emph{\BIBforeignlanguage{eng}{IEEE Sensors Journal}}, vol.~16,
  no.~1, pp. 203--209, 2016.

\bibitem{Bavirisetti2017Multi}
D.~P. Bavirisetti, ``Multi-sensor image fusion based on fourth order partial
  differential equations,'' in \emph{20th International Conference on
  Information Fusion}, 2017.

\bibitem{Zhang2017Infrared}
Y.~Zhang, L.~Zhang, X.~Bai, and L.~Zhang, ``\BIBforeignlanguage{eng}{Infrared
  and visual image fusion through infrared feature extraction and visual
  information preservation},'' \emph{\BIBforeignlanguage{eng}{Infrared Physics
  and Technology}}, vol.~83, pp. 227--237, 2017.

\bibitem{Nas}
I.~Bello, B.~Zoph, V.~Vasudevan, and Q.~Le, ``Neural optimizer search with
  reinforcement learning,'' 09 2017.

\end{thebibliography}

\end{document}